\documentclass{article}

\usepackage{PRIMEarxiv}

\usepackage[utf8]{inputenc} % allow utf-8 input
\usepackage[T1]{fontenc}    % use 8-bit T1 fonts
\usepackage{hyperref}       % hyperlinks
\usepackage{url}            % simple URL typesetting
\usepackage{booktabs}       % professional-quality tables
\usepackage{amsfonts}       % blackboard math symbols
\usepackage{nicefrac}       % compact symbols for 1/2, etc.
\usepackage{microtype}      % microtypography
\usepackage{lipsum}
\usepackage{fancyhdr}       % header
\usepackage{graphicx}       % graphics
\usepackage{amsmath}
\graphicspath{{Fig/}}
\usepackage{subfig}
\DeclareMathOperator*{\argmin}{arg\,min}

 \usepackage{rotating}
 \usepackage{amssymb}

\usepackage{dsfont}

\usepackage[style=authoryear,natbib=true,maxnames=2]{biblatex}
\title{VICatMix: variational Bayesian clustering and variable selection for discrete biomedical data}
\author{Jackie Rao\\
 {MRC Biostatistics Unit}\\
 {University of Cambridge}\\ 
 {East Forvie Building}\\
 Cambridge\\
 CB2 0SR\\
 United Kingdom\\
  \texttt{jackie.rao@mrc-bsu.cam.ac.uk} \\
   \And
 Paul D. W. Kirk \\
 {MRC Biostatistics Unit
 and CRUK Cambridge Centre Ovarian Programme}\\
 and Cambridge Institute of Therapeutic Immunology and Infectious Disease (CITIID)\\
 {University of Cambridge}\\ 
  \texttt{paul.kirk@mrc-bsu.cam.ac.uk} \\
}

\begin{document}
\maketitle

\begin{abstract}
Effective clustering of biomedical data is crucial in precision medicine, enabling accurate stratifiction of patients or samples. However, the growth in availability of high-dimensional categorical data, including `omics data, necessitates computationally efficient clustering algorithms. We present VICatMix, a variational Bayesian finite mixture model designed for the clustering of categorical data. The use of variational inference (VI) in its training allows the model to outperform competitors in term of efficiency, while maintaining high accuracy. VICatMix furthermore performs variable selection, enhancing its performance on high-dimensional, noisy data. The proposed model incorporates summarisation and model averaging to mitigate poor local optima in VI, allowing for improved estimation of the true number of clusters simultaneously with feature saliency. We demonstrate the performance of VICatMix with both simulated and real-world data, including applications to datasets from The Cancer Genome Atlas (TCGA), showing its use in cancer subtyping and driver gene discovery. We demonstrate VICatMix's utility in integrative cluster analysis with different `omics datasets, enabling the discovery of novel subtypes.\\
\textbf{Availability:} VICatMix is freely available as an R package, incorporating C++ for faster computation, at \url{https://github.com/j-ackierao/VICatMix}.\end{abstract}

% keywords can be removed

\section{Introduction}
The process of identifying groups of similar objects in data, known as \textit{cluster analysis}, is an area of research holding many key applications for biological data. For example, in precision medicine, being able to identify subtypes of disease where patients are grouped based on clinical and/or genomic data can allow clinicians to identify optimal treatments for each group and allow for stratified medicine approaches \citep{Srlie2001, Kuijjer2018, Chen2019, Verhaak2010}. Another application of clustering is in microarray analysis, where groups of samples which with similar patterns of gene expression can be identified allowing scientists to improve understanding of functional genomics \citep{Medvedovic2004, Crook2019, Dahl2007}, and further biomedical applications include, but are not limited to, areas such as identifying proteins with similar functions \citep{Zaslavsky2016} and analysing health records and surveys \citep{Molitor2010, Ni2019}. Clustering is usually performed in an unsupervised manner, and it is important that clusters are reliable in the sense that clusters can be clearly separated and differentiated, and also hold stability in other independent datasets. For example, it is important that a disease subtype can be characterised in a way so it may be clearly differentiated from other subtypes where the optimal treatment pathway could be very different.

Popular classical approaches to clustering include algorithmic methods such as k-means \citep{Hartigan1979} and hierarchical clustering \citep{Kaufman1990}, although these methods are largely heuristic and have little statistical characterisation of the resulting clusters. An alternative view for the clustering problem is by using model-based clustering, where observations are treated as samples from a mixture model distribution. Finite mixture models are mixtures of probability distributions, where each distribution corresponds to a cluster and is given a different weight pertaining to the proportion of observations assigned to the cluster \citep{McLachlan2019}. The expectation-maximisation (EM) algorithm can be used to perform maximum likelihood estimation for the model parameters \citep{Fraley2002, McLachlan2019} in a finite mixture model which has a fixed number of clusters $K$; for example, the R package \textit{mclust} implements the EM algorithm to fit a finite mixture of Gaussian distributions for model-based clustering and classification \citep{mclustref}. 

An obstacle to the use of these finite mixture models is that prior knowledge of the finite number of components $K$ in the model is needed in order to fit data to the distribution. However, the true number $K$ is not usually known, and determining this number is an important issue \citep{celeux}. Frequentist methods of circumventing this issue include performing model selection with information criterion such as the Bayesian Information Criterion (BIC) \citep{Schwarz1978, ae57af04-fd1f-3951-8417-21b68817fba1} or the integrated classification likelihood (ICL) \citep{Biernacki2000}. Both of these are employed in \textit{mclust}, as well as in \textit{FlexMix} \citep{Leisch2004}, another implementation of the EM-algorithm in R which also allows for mixtures of linear regression models, generalised linear models and other frequentist models. However, these methods require fitting many finite mixture models with a range of $K$, and may not be appropriate for the clustering problem; for example, the BIC criterion has been shown to underestimate the true number of clusters \citep{Zhao2015}. 

Alternatively, Bayesian approaches to model-based clustering can allow for the number of clusters to be a parameter for inference \citep{Miller2017, Nobile2007}. One example is the R package \textit{BayesBinMix} \citep{RJ-2017-022}, an implementation of finite mixture models for binary data, allows $K$ to be a parameter for inference by having a discrete prior over $1:K_{max}$ to determine the number of clusters in the model. Priors are also placed upon the other parameters in the model, such as the mixing weights and the cluster-specific parameters. Other Bayesian approaches use reversible jump Markov Chain Monte Carlo (RJMCMC) \citep{Richardson1997, Tadesse2005}, moving between mixture models with different values of $K$, or use `overfitted' mixture models, where $K$ is set to be more than the number of clusters expected, and a sparse Dirichlet prior on the mixing weights encourages emptying of unnecessary clusters \citep{Rousseau2011, FrhwirthSchnatter2018}. 

Bayesian approaches also allow for a range of other mixture models. A popular example is the Dirichlet process mixture model \citep{Maceachern1994, Escobar1995}, a Bayesian nonparametric model using a Dirichlet process prior to allow for an infinite number of clusters, where the number of non-empty clusters can be inferred using MCMC samplers, and the number of clusters is allowed to grow unboundedly as the number of observations increases \citep{Wade2018, Mller2013}. These models have been implemented for the clustering of biological data frequently in the literature - examples of these include an application to health survey analysis \citep{Molitor2010} and in the R package \textit{PReMiuM} \citep{Liverani2015}.

By far the most popular method for estimation of the intractable posterior in both the finite and infinite models is Markov Chain Monte Carlo (MCMC), allowing for (asymptotically) exact samples from the target density \citep{Robert2004, Bishop2006}. However, there are some major drawbacks concerning the use of MCMC. Retrieving one optimal clustering structure from the MCMC samples which is most representative of the posterior is difficult due to the multimodality of the posterior surface, and there are often mixing issues where chains are prone to being stuck, making it difficult for some MCMC samplers to reach convergence \citep{Celeux2000, Wade2018, vanHavre2015}. There is also the `label switching' phenomenon, where the same clustering structure appears to be different across the MCMC samples as the labels associated with the clusters change \citep{Redner1984}, requiring algorithms or summarisation methods to circumvent this problem. Finally, MCMC algorithms are also generally very computationally expensive; they require a large number of iterations to sufficiently explore the high-dimensional posterior surface, and often are infeasible when datasets are large.

An alternative to MCMC methods is variational inference (VI), which is a deterministic approximation method allowing us to instead turn the inference problem to an optimisation problem, by finding an approximation to the posterior. It is consistent for the estimation of mixture models under certain assumptions \citep{ChriefAbdellatif2018}, and it is much more efficient, allowing us to scale mixture models to larger datasets \citep{Blei2017}. Implementations of VI in mixture models perform well in practice on a variety of datasets including in computational biology \citep{Blei2006, Constantinopoulos2006, guan_variational_2010}. However, VI is an approximate inference method; solutions are sensitive to initialisation and often are stuck in local optima \citep{Bishop2006}. To circumvent this, we utilise a novel method of implementing a co-clustering matrix - similar to posterior similarity matrices used often in post-processing of MCMC output - to average over multiple initialisations, while still maintaining efficiency. 

Biological data, especially `omics' data is often high-dimensional where only a subset of the variables are relevant to the inherent clustering structure, motivating the use of methods which allow for variable selection. Methods \citep{Fop2018, Crook2019} for the finite mixture model case have included model selection via a greedy search algorithm over different subsets of variables \citep{Dean2009, Maugis2009}, or inclusion of latent covariate selection variables to represent feature saliency \citep{Constantinopoulos2006, White2014, Law2004}.

The availability of diverse `omics' datasets not only motivate the need for methods applicable to high-dimensional problems, but also motivates the need for methodology for integrative analysis. The improvement in performance of statistical analysis of these datasets when considering a diverse range of `omics' data together has been widely explored in the literature \citep{Kirk2012}. The Cancer Genome Atlas (TCGA) consortium have defined an abundance of cancer subtypes by integrating `omics' data including DNA methylation, gene expression, micro-RNA sequencing and somatic copy number data \citep{TCGAOV2011, CGAN2012, Colorectal2012}. 

In the next section we develop VICatMix, a Bayesian finite mixture model allowing for variable selection which incorporates variational learning to allow for a computationally efficient model. It includes variable selection and uses ideas from Bayesian model averaging to improve inference for a more stable characterisation of the clustering structure. In Sections 3 and 4 we present examples and results on both simulated and real-world data, including an integrative clustering example, and finally in Section 5 we discuss the results and future work.

\section{Methods}\label{section:methods}

\subsection{Finite mixture models with variable selection}\label{subsec:finitemix}

We model the generating distribution for the data as a mixture of a finite number $K$ of components arising from a certain parametric family, with each observation generated by one component.  The general equation for a finite mixture model with $K$ components is given by:
\begin{equation}
p(X|\boldsymbol{\pi}, \Phi)= \sum_{k=1}^K \pi_k f(X | \Phi_k),\label{mixture}
\end{equation}
where the component densities $f(X | \Phi_k)$ are from the same parametric family but with different parameters associated with each component. In our case, we model each component with a categorical distribution across $P$ random variables (which is equivalent to the Bernoulli distribution if there are two categories) and our model parameters are given by $\Phi_{kj} = [\phi_{kj1}, ..., \phi_{kjL_j}]$, where $\phi_{kjl}$ represents the probability of variable $j$ taking value $l$ on component $k$, and $L_j$ is the number of categories for variable $j$. 

In Equation~\eqref{mixture}, ${X} = \{{\bf x}_1, ..., {\bf x}_N\}$ denotes our observed data, where ${\bf x}_n$ represents one observation of $P$ categorical random variables . We denote the mixture weight associated with the $k$-th component by $\pi_k$, which satisfy $\sum_{k=1}^K \pi_k = 1$ and $\pi_k \geq 0$, and denote the parameters associated with the $k$-th component by $\Phi_k = \{\Phi_{kj}\}_{j= 1, \ldots, P}$.

\subsubsection{Variable selection}
As in \citet{Law2004} and \citet{Tadesse2005}, we introduce binary variable selection indicators $\gamma_j$ such that $\gamma_j = 1$ if and only if the $j$-th covariate is selected for inclusion in the mixture model. We now write for the probability density for a data point $x_n$ in a cluster $k$:
\begin{align}
f({\bf x}_n|\Phi_k) &= \prod_{j = 1}^P f_j(x_{nj} | \Phi_{kj})^{\gamma_j} f_j(x_{nj} | \Phi_{0j})^{1 - \gamma_j}
\end{align}

$\Phi_{0j} = [\phi_{0j1}, ..., \phi_{0jL_j}]$ represents parameter estimates obtained for covariate $j$ under the null assumption that there exists no clustering structure in the $j$-th covariate. 

\subsection{Variational Bayesian mixture models}\label{subsec:bayesian}

To complete the model specification for our Bayesian finite mixture model, we introduce priors for the model's parameters as follows:
\begin{equation}
    \boldsymbol{\pi} = (\pi_1, ..., \pi_K) \sim \mbox{Dirichlet}(\alpha_0)
\end{equation}
\begin{equation}
    \Phi_{kj} = (\phi_{kj1}, ..., \phi_{kjL_j}) \sim \mbox{Dirichlet}(\epsilon_{kj1}, ..., \epsilon_{kjL_j})
\end{equation}
\begin{equation}
\gamma_j | \delta_j \sim \mbox{Bernoulli}(\delta_j)
\end{equation}
\begin{equation}
    \delta_j \sim \mbox{Beta}(a)
\end{equation}

We set $K$ to be higher than the number of clusters we expect to find, and set $\alpha_0 < 1$. It has been shown theoretically that if $K > K_{\mbox{true}}$ (the true number of clusters), then -- under certain assumptions -- setting $\alpha_0 < 1$ in the symmetric Dirichlet prior for the mixture weight allows the true posterior to asymptotically converge to $K_{true}$ clusters as the number of observations goes to infinity, where superfluous components are emptied and the true number of clusters can be determined by counting the number of non-empty clusters \citep{Rousseau2011, vanHavre2015, MalsinerWalli2014}. This is known as an overfitted mixture model, or a sparse finite mixture, where the sparse Dirichlet prior favours mixture weights close to 0 \citep{FrhwirthSchnatter2018}.

We set $\epsilon_{kjl} = 1/L_j$ for each $k = 1, ..., K, j = 1, ..., P, l = 1, ..., L_j$, representing no prior favouring for a variable to take any certain value within any cluster. 

The hyperprior for $\delta$, the hyperparameter for the Bernoulli prior on $\gamma$, allows the prior probability of including each covariate to be a target for inference in the model. We see in experiments (Supplementary Figure \ref{cexp_ARI}) that the model's accuracy is robust to the choice of $a$, so we set $a = 2$ in all studies. When fitting the model, we set $c_j = \mathbb{E}_\gamma(\gamma_j) = 1$ initially for all $j = 1, ..., P$, so all variables are initially included and the model gradually removes variables if it does not support its inclusion.

\subsubsection{Variational inference (VI)}
We adopt a variational approach to inference, approximating the posterior parameter distribution by a distribution $q(\theta)$ that maximises the Evidence Lower Bound (ELBO), which is equivalent to minimising the Kullback-Leibler (KL) divergence between $p(\theta | X)$ and $q(\theta)$. We constrain $q$ to be a mean-field approximation, so it is a product of the form $q(\theta) = q_Z(Z)q_\pi(\pi)q_\Phi(\Phi)q_\gamma(\gamma)q_\delta(\delta)$, and $Z$ is a collection of latent variables for the data points representing their cluster allocations (see Supplementary Material). 

The variational update equations are provided in the Supplementary Material, where, to accelerate computation, we precompute estimates of the parameters $\Phi_{0j}$ associated with the covariates that do not contribute to the definition of the clustering structure, as in \citet{Savage2013}.

\begin{figure}[ht]
    \centering
    \includegraphics[scale=0.4]{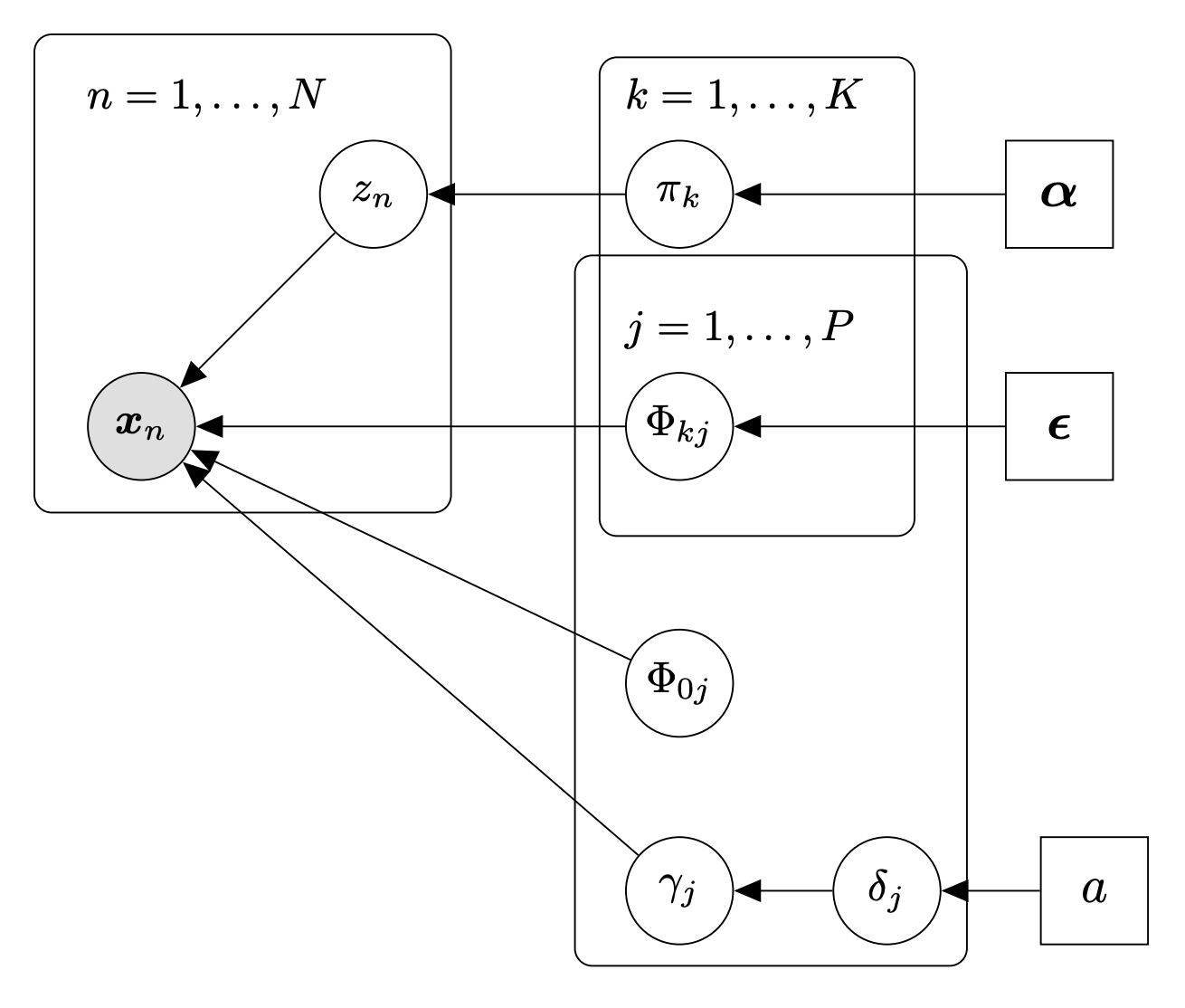}
    \caption{Graphical representation of VICatMix. Here, ${\bf z}_n$ is a `1-of-K' latent variable associated with the data point ${\bf x}_n$ representing its cluster allocation; see the Supplementary Material for more details.}
    \label{plate}
\end{figure}

Figure \ref{plate} provides a graphical representation of our variational Bayesian mixture model with variable selection.

\subsection{Summarising and Bayesian model averaging}\label{subsec:modelaverage}
Since the ELBO is a non-convex objective function, variational inference procedures can only guarantee convergence to a local optimum, which can be sensitive to initialisation \citep{Bishop2006, proximityVI}. A standard approach to addressing this challenge is to perform multiple runs of VICatMix with different random initialisations, and then to select from amongst these the run that provides the maximum ELBO \citep{Ueda2002, Friston2007}. Here we use ideas from post-processing of MCMC clustering algorithms and the SUGSVarSel approach of \citet{Crook2019} to `average' out over a range of initialisations to generate a single summary clustering ${\bf Z}^*$. This averaged model is referred to as VICatMix-Avg. In other contexts, we have found that such approaches can help to reduce the identification of spurious singleton clusters \citep{chaumenyetal}.

The central quantity we use to process and summarise $M$ clustering solutions, labelled $m = 1, ..., M$, is a $N x N$ co-clustering matrix $\textbf{P}$, defined by:
\begin{equation}
    P_{ij} = p({\bf z}_i = {\bf z}_j | X) \approx \frac{1}{M} \sum_{m = 1}^M \mathds{1} [{\bf z}_i^{(m)} = {\bf z}_j^{(m)}], \qquad i, j = 1, ..., N \label{psmeqn}
\end{equation}
where $\mathds{1}$ represents the indicator function, and ${\bf z}_i^{(m)}$ is the latent variable representing the cluster allocation for observation $i$ in clustering solution $m$. Each entry of the matrix represents an estimation of the probability that two observations will appear in the same clustering, and is analogous to the posterior similarity matrix \citep{Fritsch2009} used in post-processing of MCMC samples. We then use this quantity to find a summary representative clustering ${\bf Z}^*$. Having evaluated the co-clustering matrix, we consider two previously proposed methods for identifying a final summary clustering: `Medvedovic' clustering, where $1 - \textbf{P}$ is a distance matrix for agglomerative hierarchical clustering \citep{Medvedovic2004}, and variation of information (VoI) \citep{Meil2007,Wade2018} with optimisation methods `average' and `complete'. See also the Supplementary Material for more details on implementation.  

\subsubsection{Summarising the selected variables}
We summarise the variables selected over $M$ runs (each with a different random initialisation), by first calculating the proportion of runs in which each variable was selected, and then thresholding these proportions to identify a final summary set of selected variables. We consider thresholds $\tau = 0.5$ and $0.95$ - see Supplementary Information for full details.

\subsection{Implementation and availability}
VICatMix is implemented as an R package, where computation is accelerated using C++ (via Rcpp and RcppArmadillo). This package is freely available at \url{https://github.com/j-ackierao/VICatMix}.

\section{Examples}\label{section:examples}

\subsection{Simulation setup}\label{subsec:sims}

\subsubsection{Individual VICatMix simulations}

We generate sample binary data with $N$ observations with a given number of covariates $P$, where the probability $p$ of a `1' in each cluster for each variable is randomly generated via a $\mbox{Beta} (1, 5)$ distribution, encouraging sparse probabilities which vary across clusters. For noisy variables, the probability of a `1' is also generated by a $\mbox{Beta} (1,5)$ distribution but this probability is the same regardless of the cluster membership of the observation.  We test the effects of changing the $\alpha$ hyperparameter in our model between a range of values \{0.005, 0.01, 0.05, 0.1, 0.5, 1, 5\}. We simulate datasets with $N=1000$, $P = 100$, 10 evenly sized true clusters and initialised with $K = 30$, and look at 10 different initialisations of the model. We also test this with the same dataset size, but with 4 unevenly sized true clusters and initialise this with $K = 10$.

\subsubsection{VICatMix-Avg simulations}
In our second simulation study, we demonstrate the improvement in accuracy for VICatMix when we implement model averaging as described previously. We generate 10 independent datasets and run VICatMix or VICatMixVarSel with 30 different initialisations on each dataset. We compare the effects of using Medvedovic clustering and VoI with average and complete linkage as summarisation methods when using 5, 10, 15, 20, 25 and 30 different clustering solutions in the co-clustering matrix, and refer to the averaged models as VICatMix-Avg or VICatMixVarSel-Avg. An outline of the different simulation scenarios for this study is given in Table \ref{sims2table}.

\begin{table}[h!]
\begin{center}
\caption{Table giving parameters for data generation for the second and third simulation studies. Simulations 2.1, 2.2 and 2.3 were run without variable selection; Simulations 2.4, 2.5 were run with variable selection. $K_{init}$ refers to the initialised value of $K$ in VICatMix.}\label{sims2table}
    \begin{tabular}[c]{c | c | c | c | c | c | c }
    \centering
    ID & \% relevant & $N$ & $P$ & $K_{init}$ & $K_{true}$ & $N$ per  \\
      & variables  &   &  &  &  & cluster \\
    \hline
    2.1 &  100\% & 1000 & 100 & 30 & 10 & 50-200\\
    2.2 &  100\% & 1000 & 100 & 30 & 10 & 10-400 \\
    2.3 &  100\% & 2000 & 100 & 40 & 20 & 50-200 \\
    2.4 & 75\% & 1000 & 100 & 30 & 10 & 50-200 \\
    2.5 & 50\%  & 1000 & 100 & 30 & 10 & 50-200 \\
        \hline
    3.1 &  100\% & 1000 & 100 & 20 & 10 & 100\\
    3.2 &  100\% & 1000 & 100 & 20 & 10 & 25-400 \\
    3.3 &  100\% & 2000 & 100 & 30 & 20 & 50-200 \\
    3.4 & 75\% & 1000 & 100 & 20 & 10 & 50-200 \\
    3.5 & 50\%  & 2000 & 100 & 20 & 10 & 100-400 \\
    \end{tabular}
\end{center}
\end{table}

\subsubsection{Comparator methods}
We compare to the following existing methods: PReMiuM \citep{Liverani2015}, Bayesian Hierarchical Clustering (BHC) \citep{Heller2005, Savage2009}, BayesBinMix \citep{RJ-2017-022}, FlexMix \citep{Leisch2004} and agglomerative hierarchical clustering \citep{Kaufman1990, Langfelder2012-jq}.  Further details for the settings and implementation for them is provided in the Supplementary Material. An outline of our five simulation scenarios is captured in Table \ref{sims2table}. All models are compared to VICatMix in Simulations 3.1, 3.2 and 3.3 without variable selection; in Simulations 3.4 and 3.5, we compare VICatMixVarSel with VICatMix (without variable selection) as well as with PReMiuM, which incorporates variable selection, and BHC as a comparison to a method with no implementation of variable selection to see if its performance diminishes with noisier data.

\subsubsection{Run-times}
In our last simulation study, since the main advantage of using VI as opposed to MCMC is its computational efficiency, we run a simulation to measure the time taken to run the model. We first run the model for fixed $P=100$ and $N \in \{100, 200, 500, 1000, 2000, 5000, 10000, 20000, 50000 \}$, and then fixed $N=1000$ and $P \in \{10, 20, 50, 100, 200, 500, 1000, 2000 \}$. In all cases, we run VICatMix (without averaging) on 10 independently generated datasets with 10 true clusters each, and initialise with $K=20$ clusters. In the variable selection case, 80\% of the variables are relevant.

\subsubsection{Evaluation of results}
Across our simulations, we use the adjusted Rand Index (ARI) \citep{Rand1971, Hubert1985} to assess the quality of our clustering solutions compared to the true simulated labels, where 1 denotes an exact match to the true labels, and 0 represents a clustering solution which is no better than random allocation. We also look at comparisons of the number of clusters detected and run-times in all model comparison simulations. We also use $F_1$ scores to quantify the quality of variable selection in each variable selection simulation, which represents a harmonic mean between precision and recall \citep{Sundheim1992}.

\subsection{Yeast galactose data}\label{subsec:galactose}
We consider a subset of a dataset by \citet{Ideker2001}, consisting of gene expression data from the yeast galactose-utilisation pathway. DNA microarrays were used to measure mRNA concentrations under 20 different experimental conditions in yeast in the presence/absence of galactose and raffinose, and the experiment was replicated four times. Each column of the dataset represents one experimental condition, and the rows correspond to a subset of 205 yeast genes from the full dataset, consistent with other studies in the literature \citep{Medvedovic2004, Fritsch2009, Qin2006}. As in previous analyses, we compare our results to the classification defined by four GO (Gene Ontology) functional categories \citep{Ashburner2000}.

\subsection{Acute myeloid leukaemia data}\label{subsec:AML}
We demonstrate the performance of variable selection in VICatMix with an application to Acute Myeloid Leukaemia (AML) mutation data from the The Cancer Genome Atlas (TCGA) Research Network \citep{Hoadley2018, AML_TCGA}. Mutation data was retrieved via cBioPortal for Cancer Genomics \citep{Cerami2012, Gao2013, deBruijn2023}. Somatic mutations seen in at least two patients were considered for analysis, leaving 151 mutated genes and 185 patients in the dataset. We apply VICatMixVarSel-Avg to this binary dataset, where entry $(i, j) = 1$ if patient $i$ has a somatic mutation in gene $j$.

\subsection{Pan-cancer cluster-of-clusters analysis}
We apply VICatMix to pan-cancer data comprising 3,527 samples from 12 cancer types studied by The Cancer Genome Atlas (TCGA) Research Network in a multiplatform integrative analysis \citep{Hoadley2014}, and we demonstrate the use of VICatMix in integrative clustering methods. In the original analysis, \citeauthor{Hoadley2014} produced 5 different clustering structures of the samples based on DNA somatic copy number, DNA methylation, mRNA expression, microRNA expression, and protein expression. These sets of clusters were then integrated to produce a final clustering using Cluster-of-Clusters Analysis (COCA), an integrative clustering method first introduced by TCGA in a study to define subclasses of human breast tumours \citep{CGAN2012}, and used widely in other integrative cancer `omics studies \citep{Aure2017, Chen2016}. 

In COCA, a `Matrix of Clusters' combines various clustering structures from multiple datasets into one unified clustering result \citep{Cabassi2020}. Given M datasets, for each dataset $m = 1, ..., M$, we produce a clustering structure $c^m$ with $K^m$ clusters in each structure. $K^m$ need not be the same for all $m$. We define $K = \sum_{m = 1}^M K^m$ to be the total number of clusters. The `Matrix of Clusters' is a $K$x$N$ dimensional binary matrix, where N is the number of observations seen in all datasets, defined as \citep{Cabassi2020}:
\begin{equation}
    MOC_{k, m_k} =
    \begin{cases}
    1 & \text{for } c_n^{m} = m_k,\\
    0 & \text{otherwise, }
    \end{cases}
\end{equation}
where $c_n^{m}$ refers to the cluster assignment for observation $n$ in clustering $c^m$, and $m_k$ refers to the $k$th cluster in dataset $m$. \citeauthor{Hoadley2014} and other studies \citep{Aure2017, Chen2016} used consensus clustering \citep{Monti2003} in order to cluster the Matrix of Clusters; we apply VICatMix-Avg instead.

\section{Results}\label{section:results}

\subsection{Simulated data}\label{resultssims}

Results and analysis from the first, third and fourth simulation scenarios can be found in the Supplementary Material.

\subsubsection{VICatMix-Avg simulations}\label{averagingresults}

\begin{figure*}[ht]
    \centering
    \subfloat[]{\includegraphics[scale=0.55]{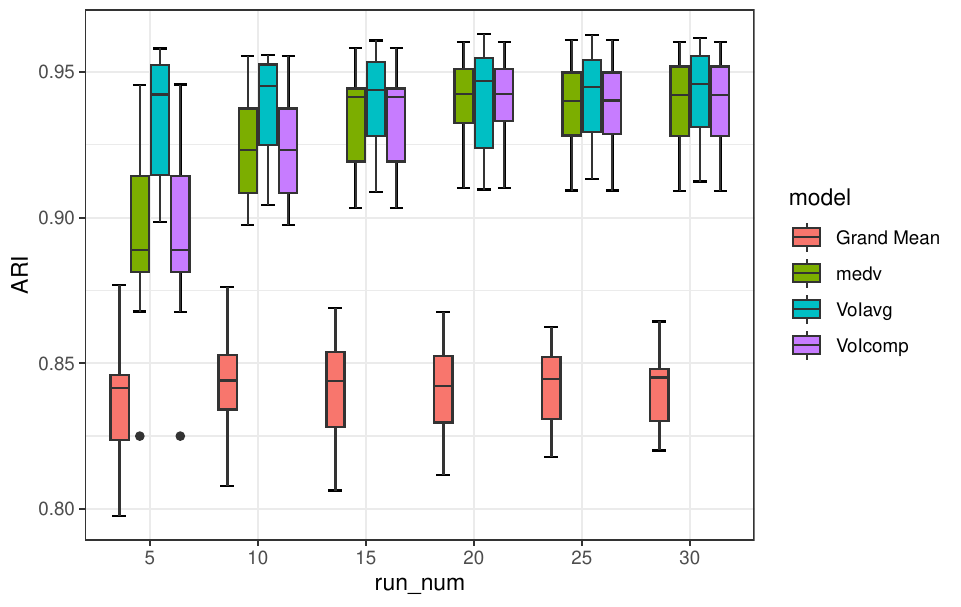}}
    \subfloat[]{\includegraphics[scale=0.55]{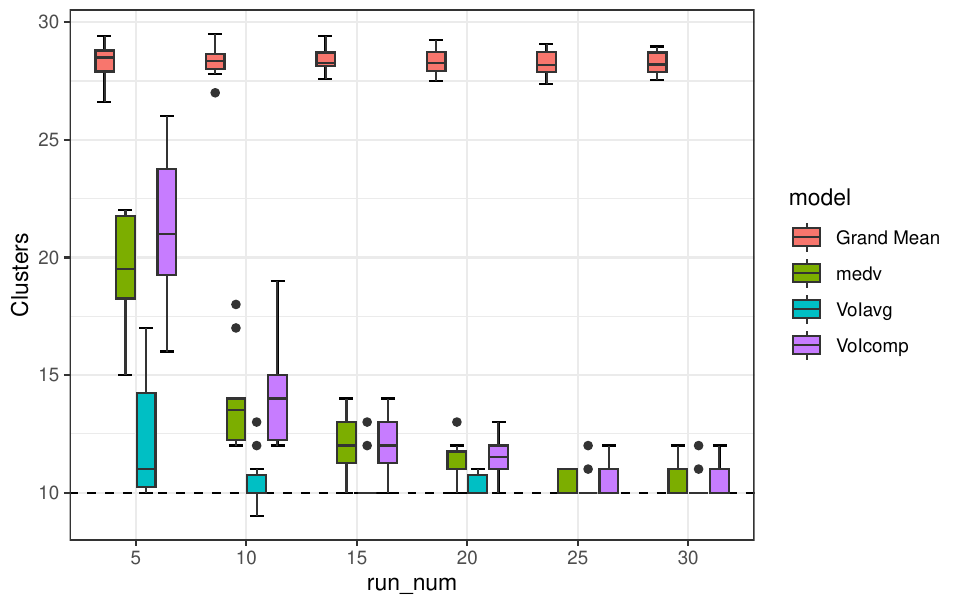}}
    \caption{Boxplots comparing the ARI and number of clusters of each model-averaging method across all 10 simulated datasets with the grand mean of the individual runs considered with different numbers of clustering solutions in the co-clustering matrix for Simulation 2.1.}
    \label{Sim1ModelAvg}
\end{figure*}

\begin{figure*}[ht]
    \centering
    \subfloat[]{\includegraphics[scale=0.55]{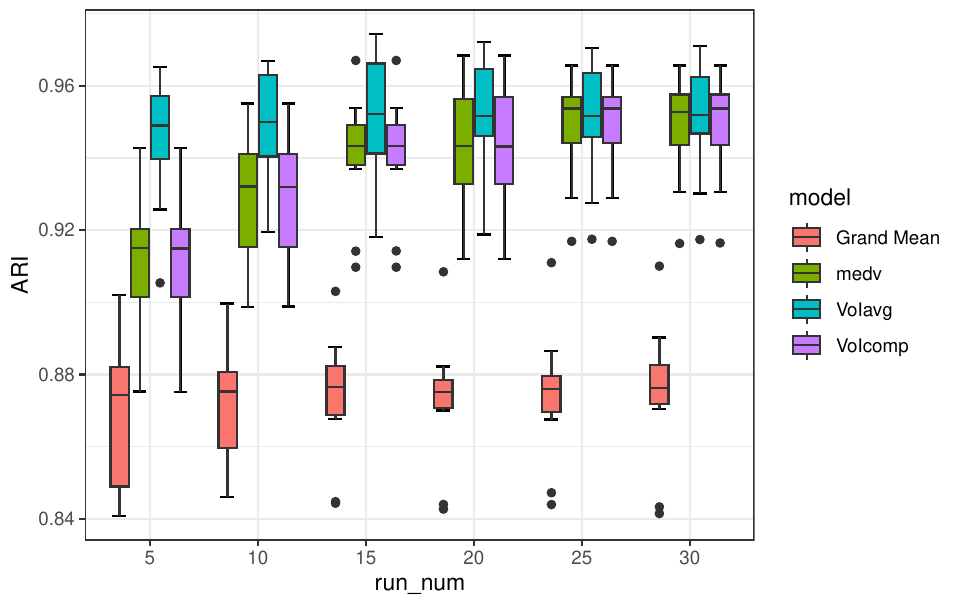}}
    \subfloat[]{\includegraphics[scale=0.55]{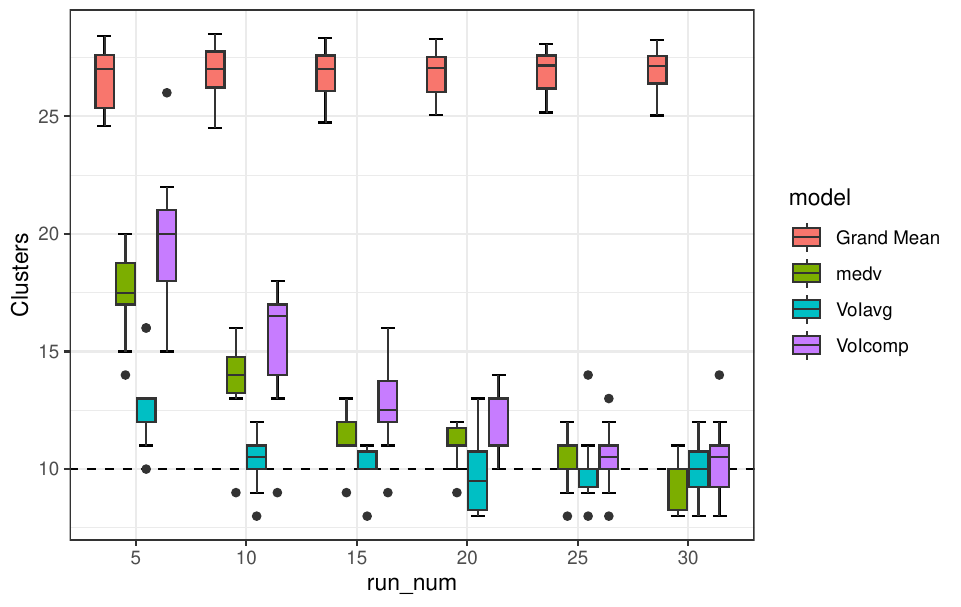}}
    \caption{Boxplots comparing the ARI and number of clusters of each model-averaging method across all 10 simulated datasets with the grand mean of the individual runs considered with different numbers of clustering solutions in the co-clustering matrix for Simulation 2.2.}
    \label{Sim2ModelAvg}
\end{figure*}

\begin{figure*}[ht]
    \centering
    \subfloat[]{\includegraphics[scale=0.55]{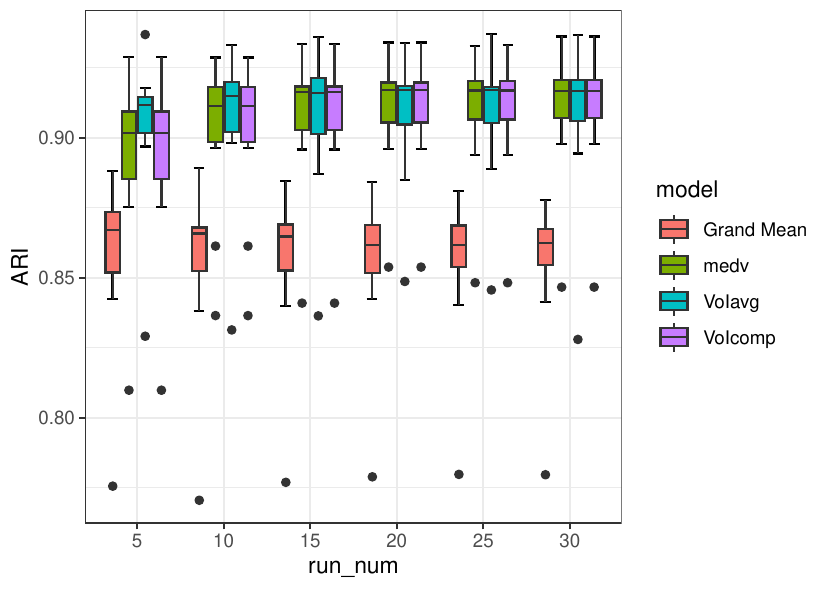}}
    \subfloat[]{\includegraphics[scale=0.55]{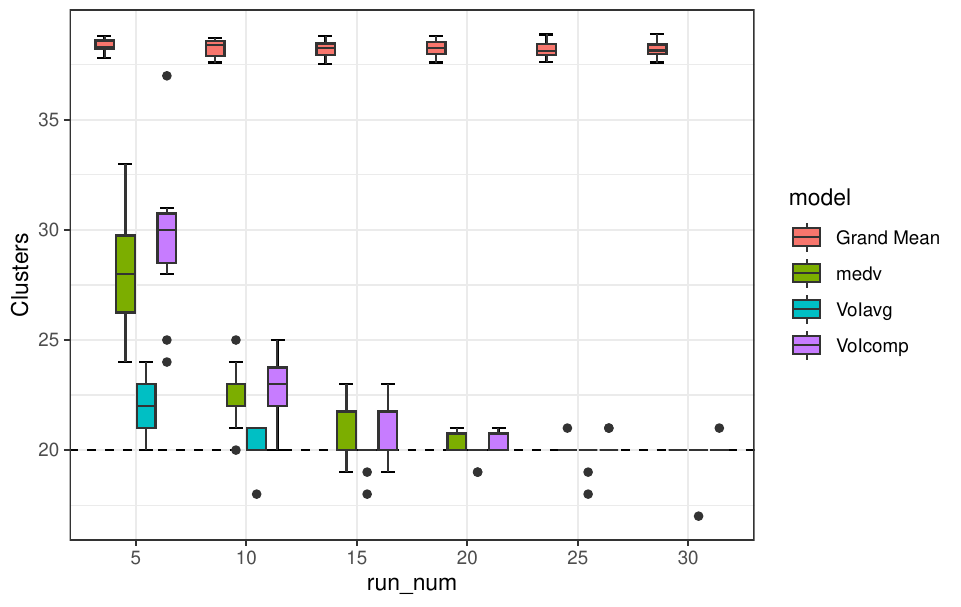}}
    \caption{Boxplots comparing the ARI and number of clusters of each model-averaging method across all 10 simulated datasets with the grand mean of the individual runs considered with different numbers of clustering solutions in the co-clustering matrix for Simulation 2.3.}
    \label{Sim3ModelAvg}
\end{figure*}

\begin{figure*}[ht]
    \centering
    \subfloat[]{\includegraphics[scale=0.25]{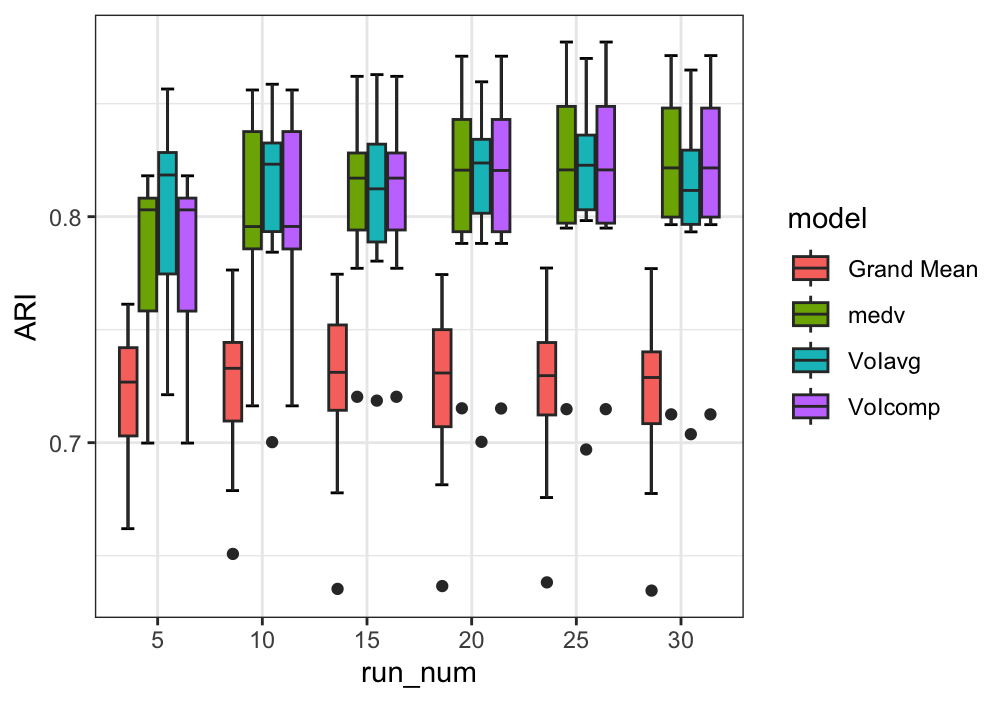}}
    \subfloat[]{\includegraphics[scale=0.25]{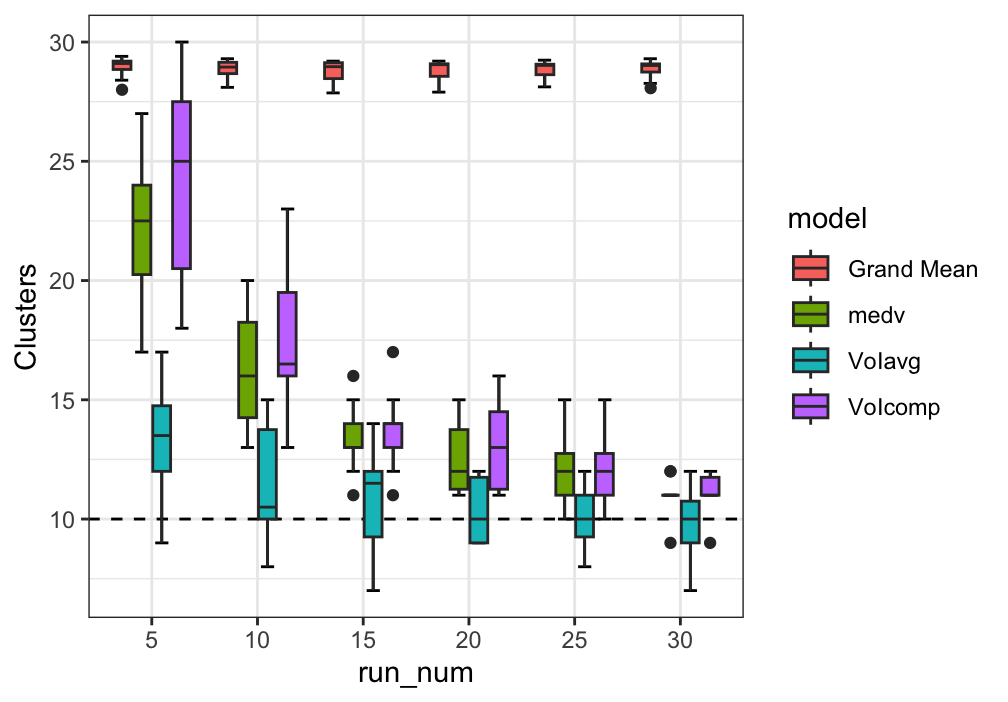}}
    \caption{Boxplots comparing the ARI and number of clusters of each model-averaging method across all 10 simulated datasets with the grand mean of the individual runs considered with different numbers of clustering solutions in the co-clustering matrix for Simulation 2.4.}
    \label{Sim1ModelAvgVar}
\end{figure*}

In Simulations 2.1 - 2.4 (Figure \ref{Sim1ModelAvg}, Supplementary Figures \ref{Sim2ModelAvg}, \ref{Sim3ModelAvg}, \ref{Sim1ModelAvgVar}, Supplementary Tables \ref{meanARISim1}, \ref{meanARISim2}, \ref{meanARIsSim3}), all summarisation methods showed a significant improvement in ARI compared to each individual run of VICatMix, even with as few as 5 runs in the co-clustering matrix. The number of non-empty clusters was much closer to $K_{true}$, suggesting that model averaging could mitigate the effect of spurious small clusters. Increasing the number of runs generally helped the averaging, and in Simulation 2.4, it is clear that the implementation of variable selection worked well to cluster noisier data with high accuracy and model averaging methods further improved this.

\begin{figure*}[ht]
    \centering
    \subfloat[]{\includegraphics[scale=0.25]{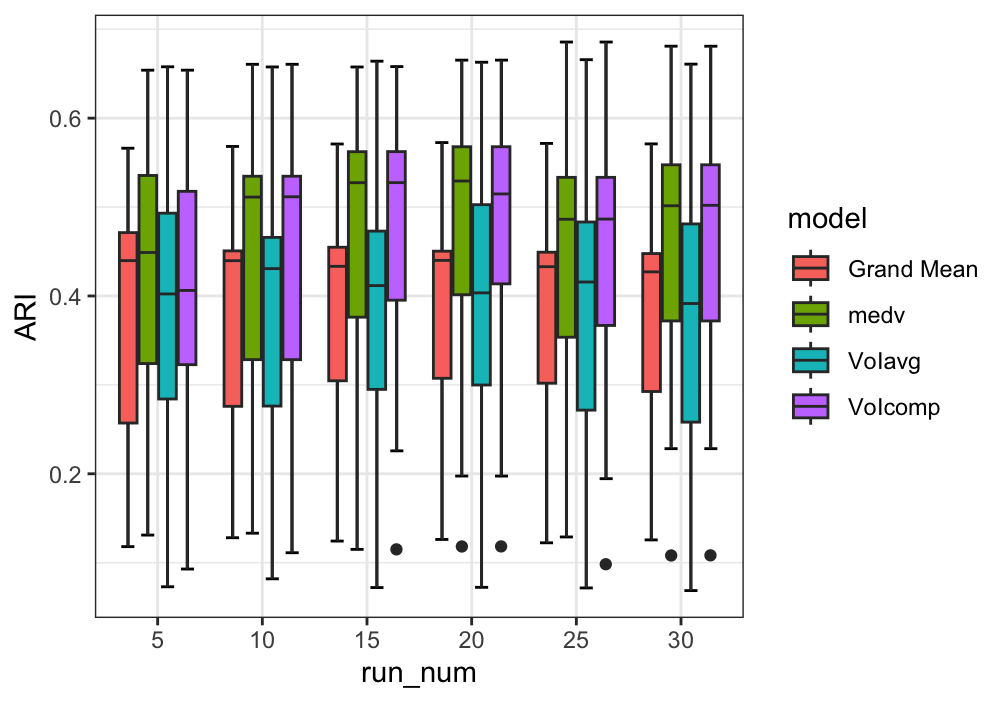}}
    \subfloat[]{\includegraphics[scale=0.25]{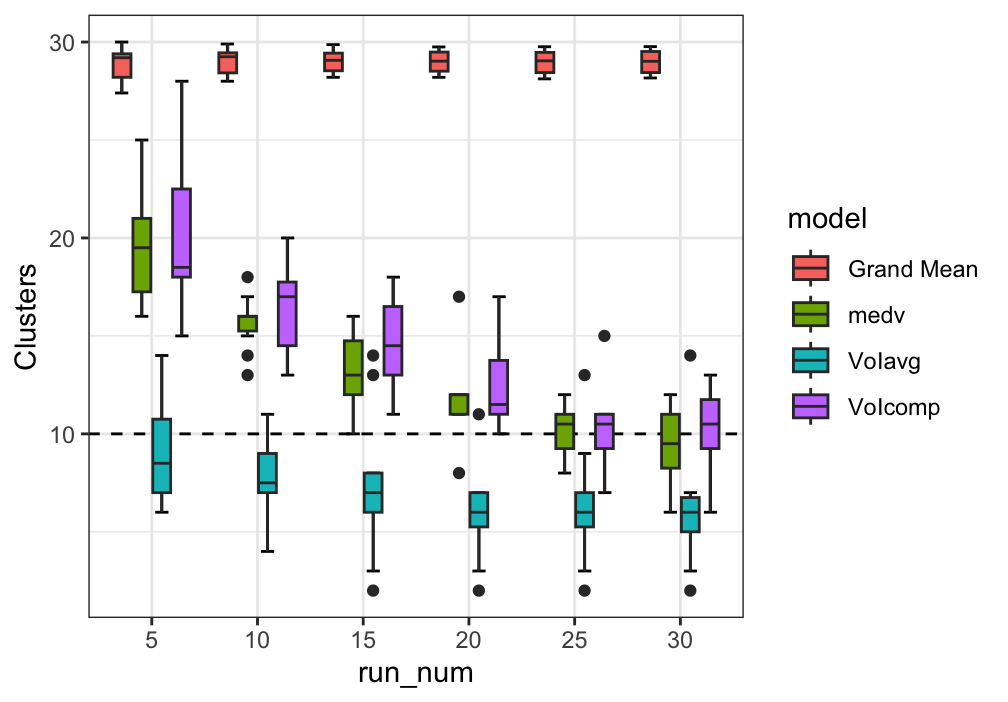}}
    \caption{Boxplots comparing the ARI and number of clusters of each model-averaging method across all 10 simulated datasets with the grand mean of the individual runs considered with different numbers of clustering solutions in the co-clustering matrix for Simulation 2.5.}
    \label{Sim2ModelAvgVar}
\end{figure*}

Simulation 2.5 (Supplementary Figure \ref{Sim2ModelAvgVar}) showed that VICatMix became less accurate when more noisy variables are added. It is unsurprising that that averaging over poorer clustering solutions did not lead to the same substantial improvement in accuracy we saw previously - particularly, VoI with average linkage tended to underestimate the true number of clusters as the number of runs was increased, and the median ARI was decreased compared to the median grand mean ARI of the individual runs in all cases. However, we saw an improvement in median ARI for Medvedovic clustering and VoI with complete linkage, and they still corrected for overestimation in the number of clusters with at least 10 individual runs in the model averaging.

In Simulations 2.4 and 2.5, we look at which variables were selected using the thresholds detailed in Section \ref{subsec:modelaverage}, and illustrate this in Supplementary Figures \ref{Sim1ModelAvgRelIrrel} and \ref{Sim2ModelAvgRelIrrel}. Table \ref{f1scores} gives mean $F_1$ scores for thresholds $\tau = 0.5$ and $\tau=0.95$ across all datasets. It is clear that the implementation of variable selection could identify the correct relevant and irrelevant variables with good accuracy, with $\tau=0.95$ seeing particularly good results in finding the correct irrelevant variables in both simulations. 

\begin{table}[ht]
\caption{Table comparing mean $F_1$ scores for variable selection methods under both Simulation 2.4 and Simulation 2.5.}
\label{f1scores}
\begin{center}
\begin{tabular}{c | c | c | c | c}
 & \multicolumn{2}{c |}{Simulation 2.4} & \multicolumn{2}{c}{Simulation 2.5} \\
\hline
Methods & 25 Runs & 30 Runs & 25 Runs & 30 Runs \\
\hline
0.5 & 0.904 & 0.905 & 0.750 & 0.747 \\
0.95 & \textbf{0.937} & \textbf{0.935} & \textbf{0.841} & \textbf{0.836} \\
\end{tabular}
\end{center}
\end{table}

\subsubsection{Implementation recommendation}

We determine that including at least 25 runs in our co-clustering matrix is sufficient, and that variation of information with complete linkage and a threshold of 0.95 for variable selection are the optimal methods to be used for VICatMix-Avg. These are the default values in the VICatMix-Avg in the R package. Each run of VICatMix can be run in parallel, given enough compute, and the calculation of an optimal clustering under VoIcomp took around a second in all our simulations. 

\subsection{Yeast galactose data}\label{resultsgalactose}
We ran VICatMix-Avg with $K = 4$, incorporating the prior assumption that there are 4 GO functional categories for the data. A heatmap visualising the results is in Supplementary Figure \ref{Galactose4Heat}. We saw that VICatMix-Avg only gave 3 clusters, where GO categories 1 and 3 remained almost intact, and the final cluster was a mix of the two smaller functional groups. This clustering structure gave us an ARI score of 0.933 with the four functional groups, comparable with ARI scores using models such as BHC and agglomerative hierarchical clustering for the same dataset in the literature \citep{Savage2009, Yeung2003}.

\begin{figure}[t]
    \centering
    \includegraphics[scale=0.23]{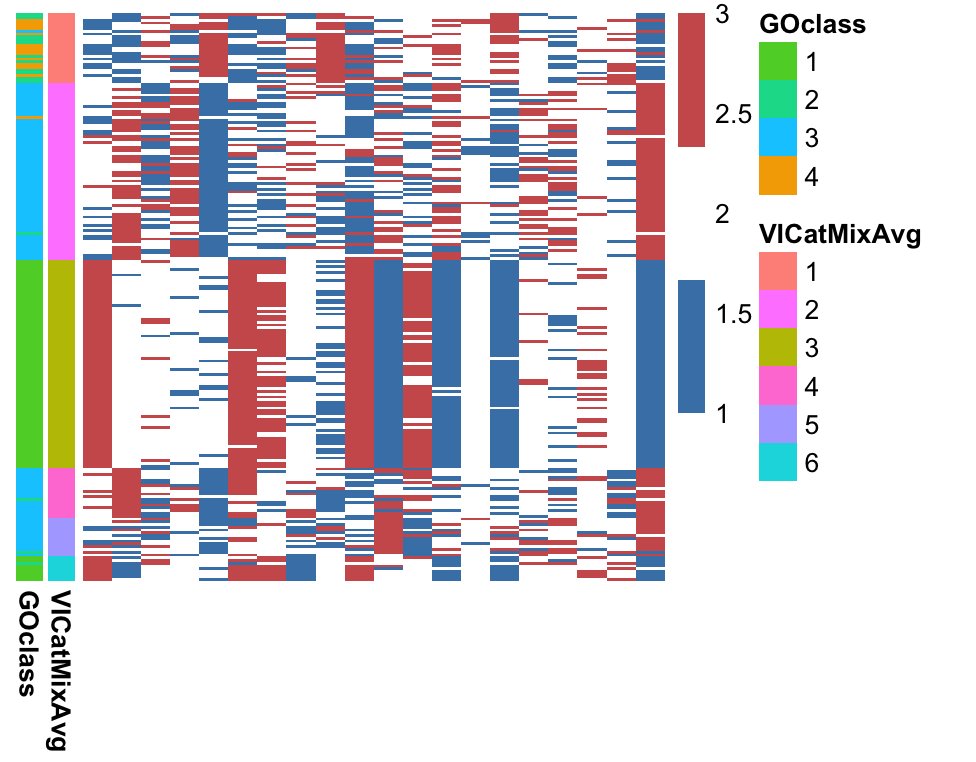}
    \caption{Heatmap of the VICatMix-Avg clustering structure on the yeast galactose data compared with the GO labelling when K=10.}
    \label{Galactose10Heat}
\end{figure}

Motivated by the ability of VICatMix-Avg to automatically detect the true number of clusters in a dataset, we also ran VICatMix-Avg with $K=10$, which also gave a clearly coherent clustering structure with 6 clusters in Figure \ref{Galactose10Heat}. We still saw a cluster which was a mix of the smaller GO functional categories, but GO category 3 was almost perfectly subdivided into three groups, which could infer the existence of biologically significant subcategories within the GO functional classes. 

\subsection{Acute myeloid leukaemia data}\label{resultsaml}

\begin{figure*}[ht]
    \centering
    \subfloat[All variables shown]{\includegraphics[scale=0.18]{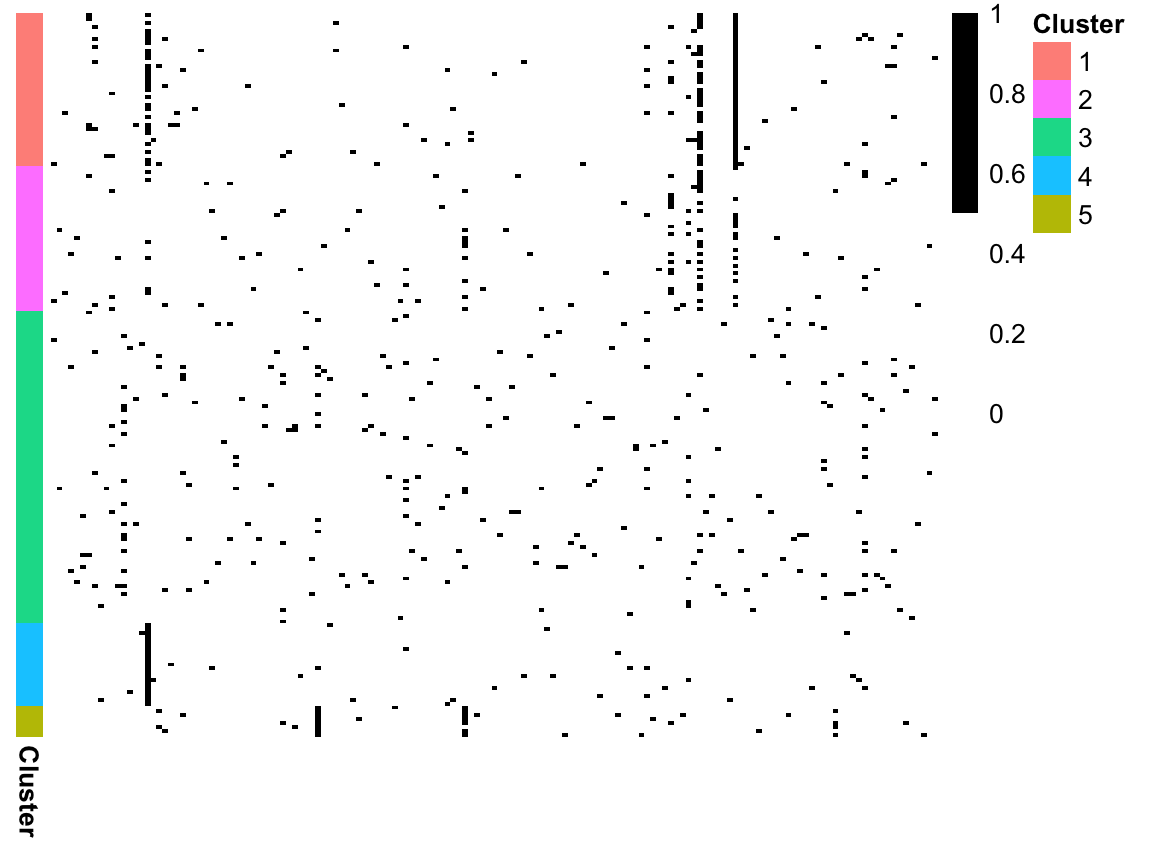}}
    \subfloat[6 selected variables shown]{\includegraphics[scale=0.18]{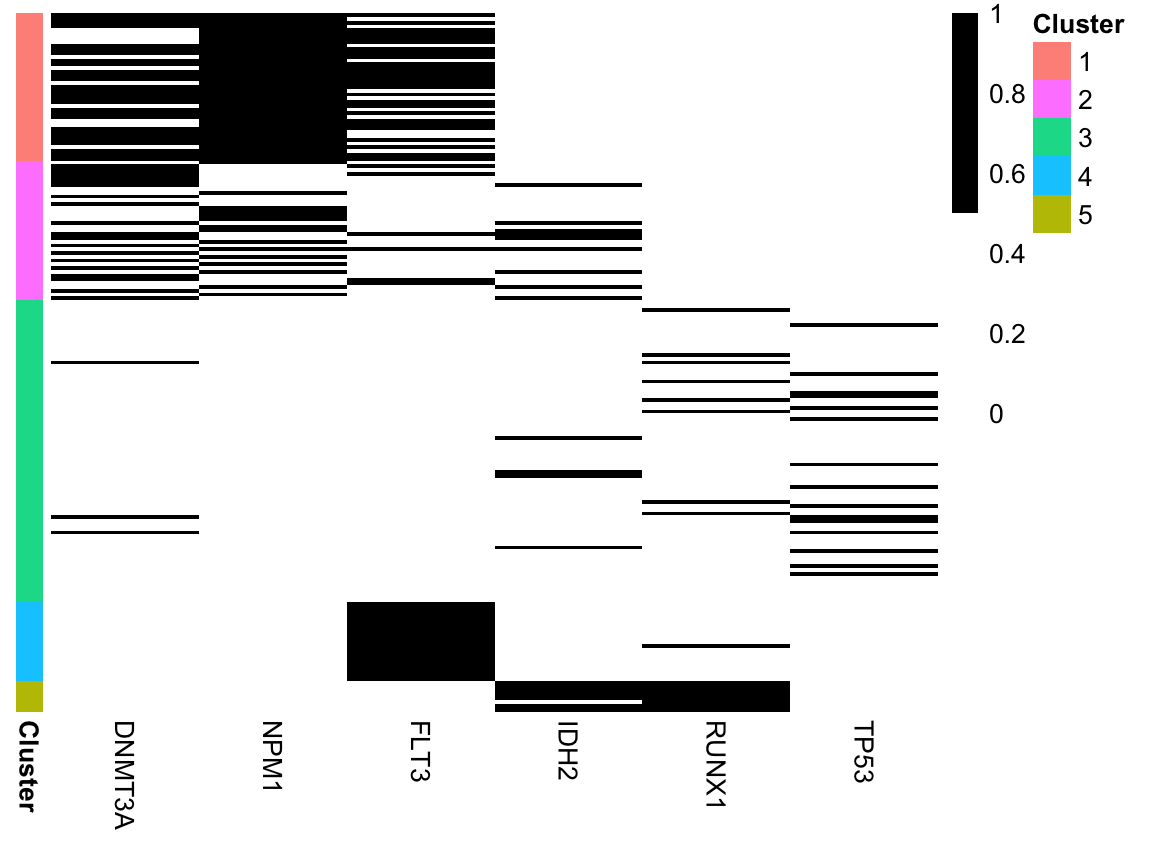}}
    \caption{Heatmaps of the VICatMixVarSel-Avg clustering structure on the AML mutation dataset.}
    \label{AMLheat}
\end{figure*}

Applying VICatMix without variable selection led to all samples consistently being put into one cluster, demonstrating the need for variable selection methods for the clustering of data that can be very noisy or sparse. Figure \ref{AMLheat} depicts heatmaps of the clustering structure generated by VICatMixVarSel-Avg with 30 initialisations, where we found 6/151 genes to be selected in more than 95\% of the runs.

\begin{figure}[ht]
    \centering
    \includegraphics[scale=0.22]{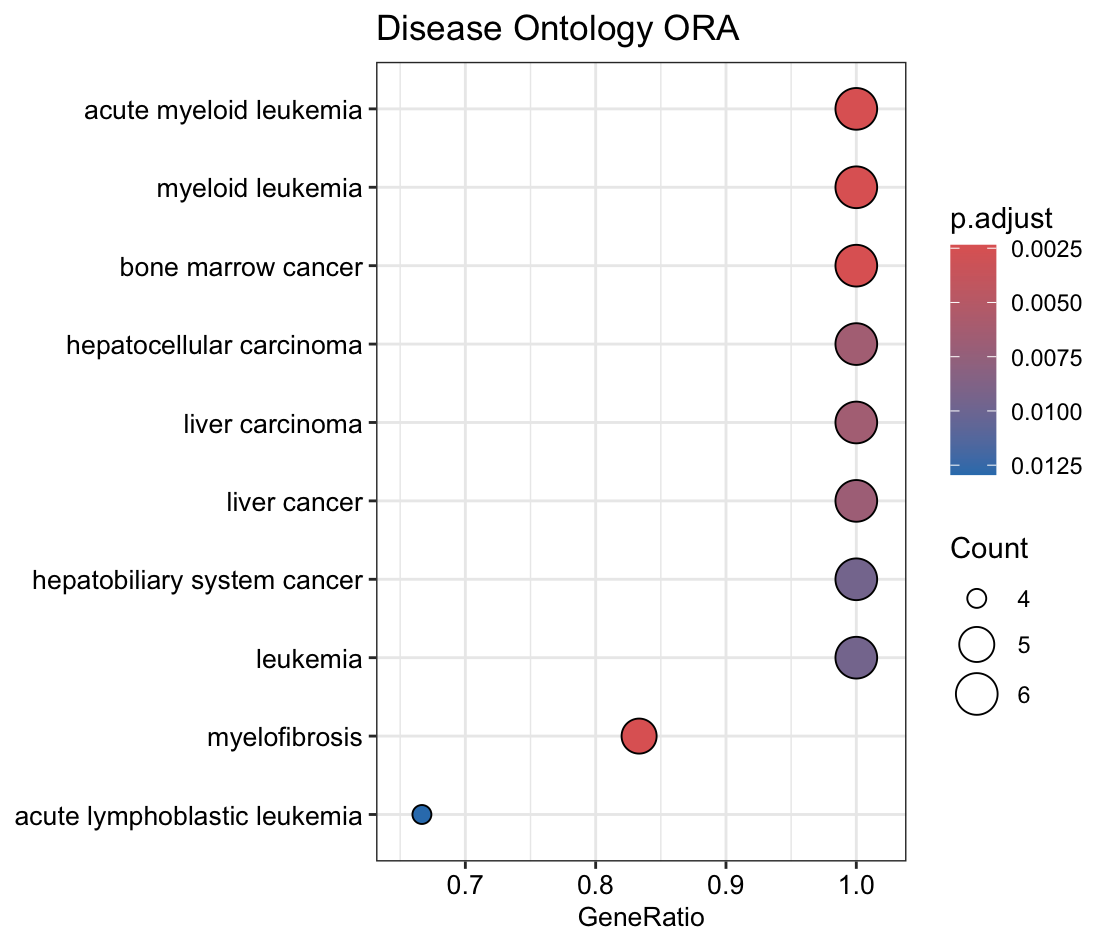}
    \caption{Dotplots visualising over-representation analysis (ORA) for 6 selected genes for the AML dataset using gene-disease annotations from the Disease Ontology (DO). p.adjust is the p-value from the hypergeometric test used in ORA, adjusted using the Benjamini-Hochberg procedure. The top 10 most significant annotations are shown.}
    \label{DO_dot}
\end{figure}

\begin{figure*}[ht]
    \centering
    \includegraphics[scale=0.22]{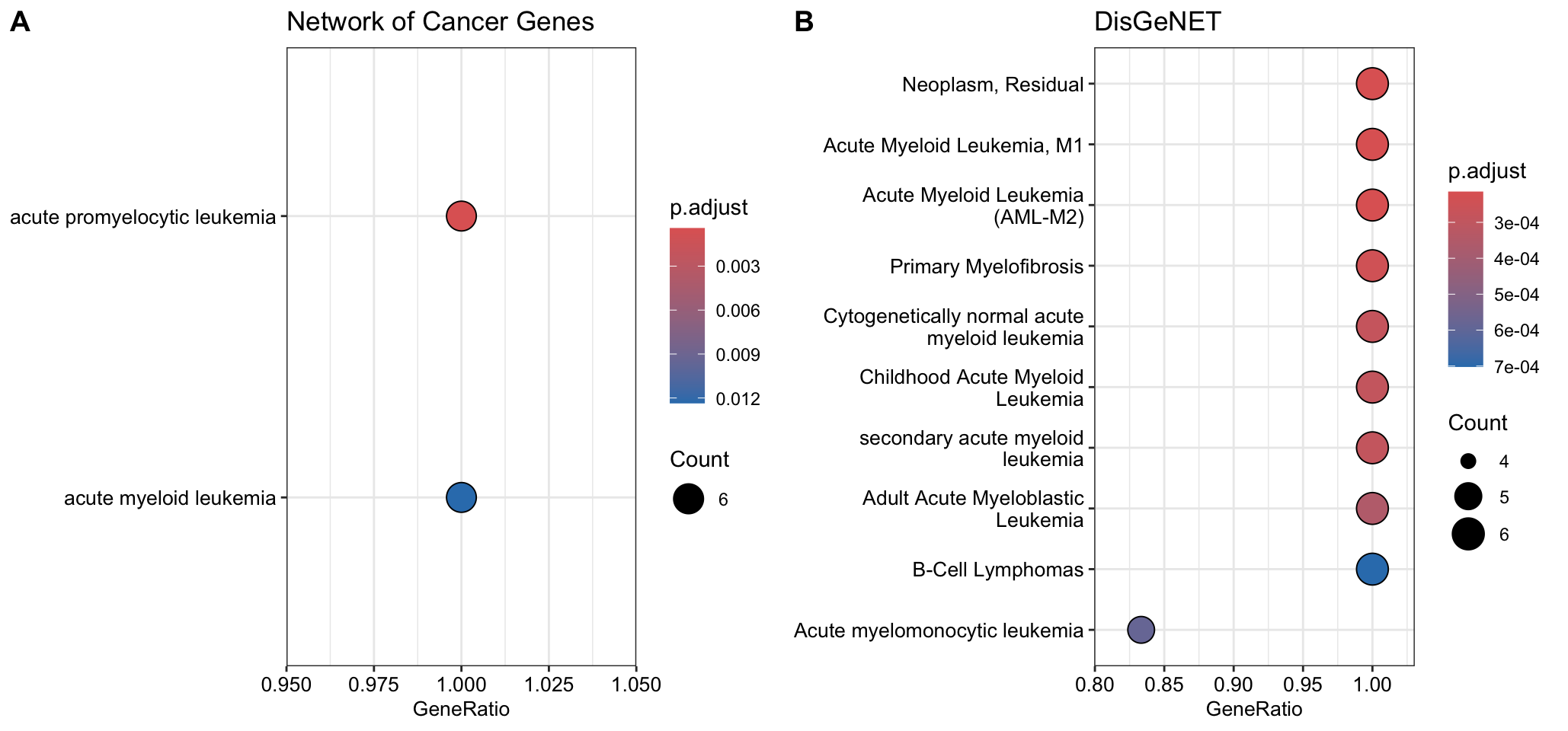}
    \caption{Dotplots visualising over-representation analysis for 6 selected genes for the AML dataset using gene-disease annotations from the Network of Cancer Genes and DisGeNET.}
    \label{NCGDGN_dot}
\end{figure*}

We performed overrepresentation analysis (ORA) for the 6 selected genes - DNMT3A, NPM1, FLT3, IDH2, RUNX1, TP53 \citep{Boyle2004} using R/Bioconductor packages ClusterProfiler \citep{Wu2021} and DOSE \citep{Yu2014}. When we performed ORA using the Disease Ontology (DO) \citep{Schriml2011} (Figure \ref{DO_dot}), we saw that the most significantly overrepresented disease annotations in our gene set were all clearly relevant for AML. DOSE also supports ORA using Network of Cancer Genes \citep{Dressler2022} and DisGeNET \citep{Piero2019}. Supplementary Figure \ref{NCGDGN_dot} demonstrates that ORA using these repositories further substantiates the association between the 6 selected genes and AML.

Recurrent mutations of all 6 genes have been associated with therapeutic and prognostic implications in AML in the literature and have played a part in molecular classification of AML \citep{Shin2016, DiNardo2016, Yu2020}. For example, all 6 genes are noted to be significantly mutated in \citet{Shin2016}, where a mutation in DNMT3A was shown to be significantly associated with adverse outcome in addition to conventional risk stratification such as the European LeukemiaNet (ELN) classification.

Additionally, the samples in Cluster 2 had \textit{either} a DNMT3A mutation \textit{or} a NPM1 mutation; this is illustrated in the Supplementary Material and could be indicative of further subclustering structure within the cluster.

\subsection{Pan-cancer cluster-of-clusters analysis}\label{resultsCOCA}

\begin{figure}[ht]
\centering
\includegraphics[scale=0.18]{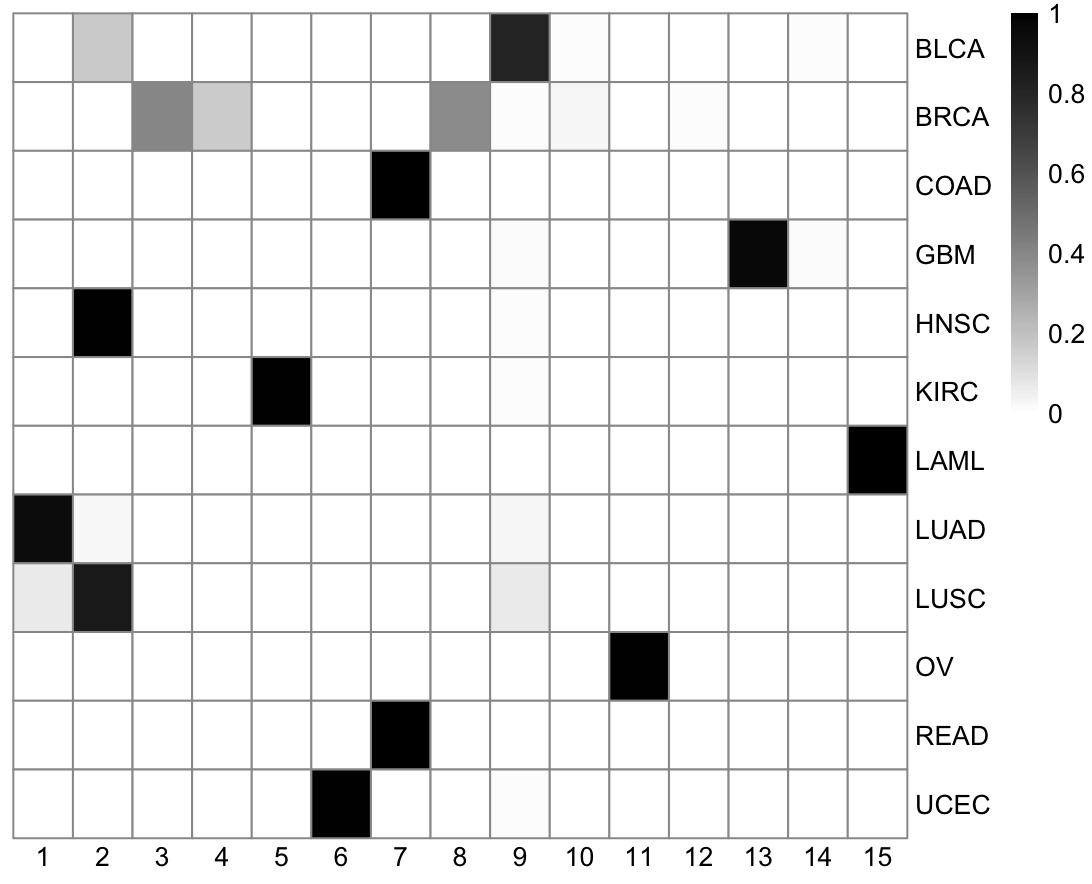}
\caption{A heatmap showing the correspondence between clusters produced by VICatMix-Avg and the tissues of origin, where $K=15$. A darker cell colour in row i indicates a higher percentage of samples from tissue i are in the given cluster j.}
\label{PercentCOCA15}
\end{figure}

\begin{sidewaysfigure*}
\centering
\includegraphics[scale=0.85]{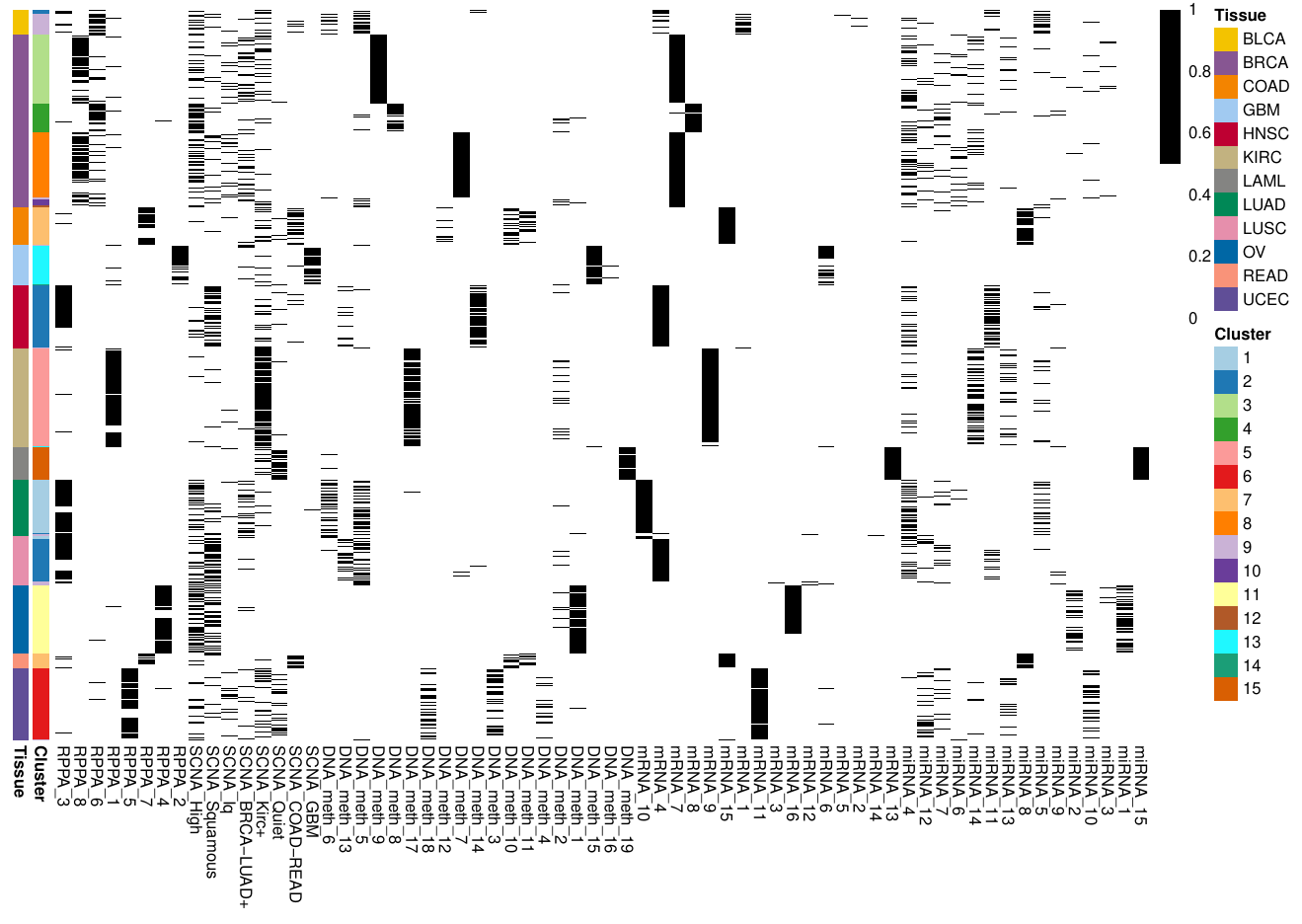}
\caption{A heatmap depicting the VICatMix-Avg clustering of the Matrix of Clusters for our pan-cancer data, where $K=15$, and we compare the clusters to the tissue of origin.}
\label{PheatmapCOCA15}
\end{sidewaysfigure*}

We applied VICatMix to the same `Matrix of Clusters' that \citeauthor{Hoadley2014} created. We ran VICatMix-Avg with 25 initialisations with no variable selection as we want to incorporate all clustering information. We initially set the upper bound on the number of clusters to be $K= 15$ to enable a similar analysis to the 11 subtypes found by \citeauthor{Hoadley2014}. We found the resulting clusters to correspond to the tissue of origin for the tumour samples as shown in Figures \ref{PheatmapCOCA15} and \ref{PercentCOCA15}. For example, LAML samples correspond perfectly with Cluster 15 and Cluster 11 is almost precisely made up of OV samples. We saw `squamous-like' cluster in Cluster 2 of mostly HNSC and LUSC samples, with other LUSC samples mostly mixing with either LUAD samples in Cluster 1 or BLCA samples in Cluster 9, which is similar to analysis by \citeauthor{Hoadley2014}. Cluster 7 consisted of a shared COAD/READ cluster, which \citeauthor{Hoadley2014} also found.

\begin{figure*}
\centering
\includegraphics[scale=0.76]{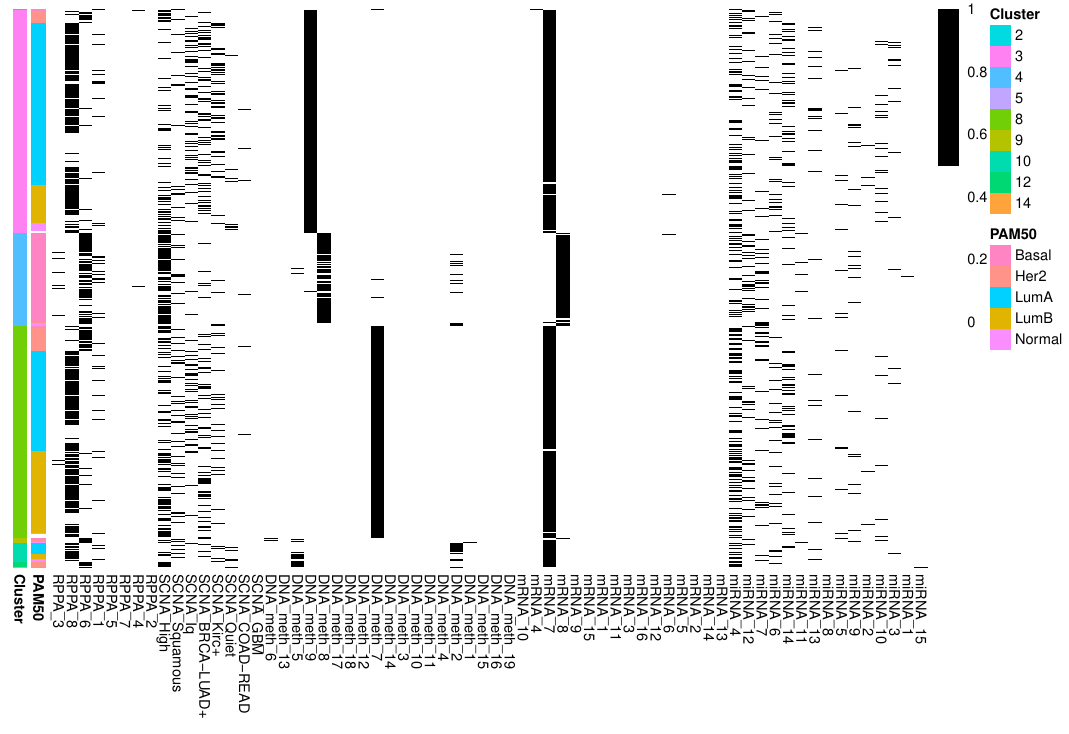}
\caption{A heatmap showing the VICatMix-Avg clustering of the Matrix of Clusters for the BRCA samples in our pan-cancer data, comparing the clusters to known PAM50 subtypes.}
\label{BRCAHeatmap15}
\end{figure*}

Focusing on the subclusters of breast cancer, we compared our clusters to PAM50 classificiations \citep{Parker2009, Perou2000, Srlie2001} by TCGA \citep{Berger2018} (Supplementary Figure \ref{BRCAHeatmap15}. We found that Cluster 4 (which almost completely coincides with the BRCA-Basal cluster of \citeauthor{Hoadley2014}) was made up of primarily Basal samples, with 132/141 Basal samples falling into this cluster, and this cluster is clearly separated from other BRCA samples. 

VICatMix-Avg's identification of the Basal BRCA subtype, which has been shown to have a statistically significantly different response to chemotherapy and prognostic outlook \citep{Parker2009, Banerjee2006}, motivates its application to the identification of other cancer subtypes in integrative `omics data analysis. To investigate further subclustering structure, corresponding to putative tumour subtypes, we additionally considered $K = 40$. The resulting clusters, their correspondence with tissue of origin, and a discussion of these results is provided in the Supplementary Material (Section \ref{suppPancan}, Figures \ref{PheatmapCOCA} and \ref{PercentCOCA}).  

\section{Discussion}\label{section:discussion}

In this paper, we have presented VICatMix, a variational Bayesian model training a finite mixture model for categorical data including variable selection. It is clear that VI provides a substantial speed-up where MCMC methods have faltered in the past due to their high computational cost. 

We demonstrated that our addition of summarisation and Bayesian model averaging via a co-clustering matrix mitigates the problems of reaching poor local optima when using VI and allows for very good results in simulations. VICatMix-Avg is both faster than many competitors in the literature and also more accurate in terms of the ARI and finding the true number of clusters.

Applications to real-world biomedical data showed that VICatMix was able to identify biologically relevant clusters. VICatMix found clusters aligning with GO functional categories in the case of yeast galactose data, and a TCGA pan-cancer dataset showed that VICatMix was able to separate most cancers by tissue type and further subtype these in an integrative clustering analysis. It would be interesting to follow the entire COCA workflow with our own clustering analysis for each type of `omics data, which would require tailoring VICatMix to be applicable to other data types including continuous data, and compare the results we get with the new Matrix of Clusters. We could further apply VICatMix to other integrative clustering methods for `omics data, such as by adapting MDI \citep{Kirk2012} or BCC \citep{Lock2013}, both of which simultaneously integrate different `omics data and perform the clustering using Dirichlet mixture models. The power of variable selection was seen in the application to AML data, where VICatMixVarSel was able to pick out 6 statistically significantly relevant genes for AML out of a total of 151.

Overall, VICatMix allows us to bypass the hurdle of computationally expensive MCMC models for cluster analysis of categorical data using VI, and while our resulting model is still only an approximation to the true posterior, our results on both simulated and real-world data show remarkable results. Further areas for improvement in the model include implementing cluster-specific regression models to investigate the relationship between covariates and a given response variable, incorporating the ELBO to give different weights for different initialisations in the co-clustering matrix for summarisation, including `merge-delete' moves to improve efficiency \citep{pmlr-v38-hughes15}, and extending to a multi-view model where multiple clustering `views' can be modelled in high-dimensional biomedical data. 

\section{Competing interests}
No competing interest is declared.

% \section{Author contributions statement}

% Must include all authors, identified by initials, for example:
% S.R. and D.A. conceived the experiment(s),  S.R. conducted the experiment(s), S.R. and D.A. analysed the results.  S.R. and D.A. wrote and reviewed the manuscript.

\section{Acknowledgments}
Kirk and Rao both acknowledge core MRC (UKRI) funding through Kirk's programme at the MRC Biostatistics Unit.   Kirk additionally acknowledges \verb+MR/S027602/1+.  This research was supported by the NIHR Cambridge Biomedical Research Centre (NIHR203312). The views expressed are those of the author(s) and not necessarily those of the NIHR or the Department of Health and Social Care.

%USE THE BELOW OPTIONS IN CASE YOU NEED AUTHOR YEAR FORMAT.
%\bibliographystyle{apacite}

%\bibliography{reference}
\printbibliography

%TC:ignore

\section*{Appendices}

\subsection{Mixture models}

We introduce latent variables $z_n$ associated with each data point $x_n$ which is a `1-of-K' binary vector in $\mathbb{R}^K$ with exactly one non-zero element; $z_{nk} = 1$ if and only if $x_n$ is associated with the $k$-th component. This allows us to ease computation for the variational algorithm. We now rewrite Equation \eqref{mixture} as:
\begin{align}
p(X|Z, \pi, \Phi) &= \prod_{n=1}^N\prod_{k=1}^K f({\bf x}_n | \Phi_{k})^{z_{nk}}
\end{align}

We also write down the conditional distribution of the latent variables $Z$ given the mixing coefficients $\pi$:
\begin{equation}
p(Z|\pi) = \prod_{n=1}^N\prod_{k=1}^K \pi_k^{z_{nk}};
\end{equation}

The overall conditional distribution for the observed data, given the latent variables and component parameters, in the case of variable selection, is given by:
\begin{align}
p(X|Z, \pi, \Phi, \gamma) &= \prod_{n=1}^N\prod_{k=1}^K (\prod_{j = 1}^P \phi_{kjx_{nj}}^{\gamma_j} \phi_{0jx_{nj}}^{1 - \gamma_j})^{z_{nk}} \label{equationwithZ}
\end{align}

\subsection{Variational inference}\label{subsec:VIsupp}
In variational inference (VI), we approximate $p(\theta | X)$ with a tractable distribution $q(\theta)$. We have that the following holds for any arbitrary distribution $q(\theta)$:
\begin{align}
    \ln p(X) = \mathcal{L}(q) + KL(q||p),
\end{align} 
where 
\begin{align}
    \mathcal{L}(q) &= \int q(\theta) \ln \left(\frac{p(X, \theta)}{q(\theta)} \right)d\theta\\
    KL(q||p) &= -\int q(\theta) \ln \left(\frac{p(\theta|X)}{q(\theta)} \right)d\theta.
\end{align} 

$KL(q ||p)$, the Kullback-Leibler (KL) divergence between $p(\theta | X)$ and $q(\theta)$, is always non-negative, so we see that $\mathcal{L}(q)$, often known as the ELBO (Evidence Lower Bound), is a lower bound for the (log) marginal likelihood, with equality if and only if $p(\theta | X)=q(\theta)$. Therefore, if we maximise $\mathcal{L}(q)$ with respect to $q(\theta)$, we minimise the KL divergence between $p(\theta | X)$ and $q(\theta)$. We can adopt an iterative procedure to optimise $\mathcal{L}(q)$, the ELBO, analogous to the EM-algorithm, and cycle between optimising $q(\theta)$ with respect to each parameter in order for our variational approximation to be as close to the targeted posterior distribution. At each iteration, we can calculate the value of the ELBO function and monitor this value for convergence.

We constrain $q$ to be a mean-field approximation, so it is a product of the form $q(\theta) = q_Z(Z)q_\pi(\pi)q_\Phi(\Phi)q_\gamma(\gamma)q_\delta(\delta)$. By rewriting $\mathcal{L}(q)$ with $q(\theta)$ in this form, the optimal solution $q_j^\ast(\theta_j)$ for each component of $\theta$, $\theta_j$, satisfies:
\begin{align}
q_j^\ast(\theta_j) &= \frac{\exp\left( \mathbb{E}_{i\ne j}[\ln p(X, \theta)]\right)}{\int \exp\left( \mathbb{E}_{i\ne j}[\ln p(X, \theta)]\right)d\theta_j} \\
\ln q_j^\ast(\theta_j) &= \mathbb{E}_{i\ne j}[\ln p(X, \theta)] + k,\label{key}
\end{align}
where $k$ is an arbitrary constant ensuring that the density integrates to 1. 

\subsection{Variational updates}

The overall conditional distribution for the observed data without variable selection can be recovered from Equation \eqref{equationwithZ} by setting $\gamma_j = 1$ for all $j$. We decompose the full model over all of our observations, latent variables and parameters including their priors as:
\begin{equation}
    p(X,Z,\pi,\Phi) = p(X|Z, \Phi)p(Z|\pi)p(\pi)p(\Phi)
\end{equation}
In this case, the variational update equations for the cluster allocation latent variables, $Z$, are given by:
\begin{equation}
q^\ast(Z) = \prod_{n=1}^N\prod_{k=1}^K r_{nk}^{z_{nk}}, \qquad r_{nk} = \frac{\rho_{nk}}{\sum_{j = 1}^K \rho_{nj}} \label{respexplanation}
\end{equation}
\begin{equation}
    \ln \rho_{nk} = \mathbb{E}_{ \pi}[\ln {\pi}_k] + \sum_{i=1}^P \mathbb{E}_{\Phi}[\ln \phi_{kix_{ni}}] \label{rhodef}
\end{equation}

We call $r_{nk}$ the responsibility of the $k$-th component for the $n$-th observation, and $\mathbb{E} [z_{nk}] = r_{nk}$; data point $n$ is allocated to the cluster $k$ with the highest responsibility.

The variational update equations for $\pi, \phi$ are given by:
\begin{equation}
    q^\ast(\pi) \propto \prod_{k=1}^K {\pi_k}^{\sum_{n=1}^N r_{nk} + \alpha_k - 1} \label{pidef1}
\end{equation} 
\begin{equation}
    \pi = (\pi_1, ..., \pi_K) \sim \mbox{Dirichlet}(\alpha^\ast_1, ..., \alpha^\ast_K) \label{pidef2}
\end{equation}
where $\alpha^\ast_k = \alpha_k + \sum_{n=1}^N r_{nk}$ for $k = 1, ..., K$.

\begin{equation}
    q^\ast(\phi_{ki}) \propto \prod_{l=1}^{L_i} \phi_{kil}^{(\epsilon_{kil} + \tilde{N_{il}} - 1)} \label{phidef1}
\end{equation}
\begin{equation}
    \phi_{ki} = (\phi_{ki1}, ..., \phi_{kiL_i}) \sim \mbox{Dirichlet}(\epsilon^\ast_{ki1}, ..., \epsilon^\ast_{kiL_i}) \label{phidef2}
\end{equation}
where we let $\tilde{N_{il}} = \sum_{n=1}^N \mathbb{I}(x_{ni} = l)r_{nk}$, and $\epsilon^\ast_{kil} = \epsilon_{kil} + \tilde{N_{il}}$ for all $l = 1, ...., L_i$, $k = 1, ..., K$ and $i = 1, ..., P$..

With variable selection, the full model is now given by:
\begin{equation}
    p(X,Z,\pi,\Phi, \gamma, \delta) = p(X|Z, \Phi, \gamma)p(Z|\pi)p(\pi)p(\Phi)p(\gamma | \delta)p(\delta)
\end{equation}

The form of the Z update remains the same as in Equation \eqref{respexplanation} but we have that $\rho_{nk}$ is defined instead by:
\begin{align}
    \ln \rho_{nk} &= \mathbb{E}_{ \pi}[\ln {\pi}_k] + \sum_{i=1}^P c_i \mathbb{E}_{\Phi}[\ln \phi_{kix_{ni}}] + (1 - c_i)(\ln \phi_{0ix_{ni}}) \label{rhodef2}
\end{align}

$c_i = \mathbb{E}_\gamma(\gamma_i)$, where the expectation is taken over the variational distribution for $\gamma$.

The form of the $\pi$ update remains the same as in Equations \eqref{pidef1}, \eqref{pidef2}. The variational update for $\phi$ also remains in the same form as in Equations \eqref{phidef1}, \eqref{phidef2}, but we redefine $\tilde{N_{il}} = \sum_{n=1}^N \mathbb{I}(x_{ni} = l)r_{nk}c_i$.

The updates for our variable selection parameters $\gamma$ and $\delta$ are given as:
\begin{equation}
    q^\ast(\gamma_i) = c_i^{\gamma_i} (1 - c_i)^{1-\gamma_i}, \qquad c_i = \frac{\eta_{1i}}{\eta_{1i} + \eta_{2i}} = \mathbb{E}_\gamma(\gamma_i)
\end{equation}
\begin{align}
    \ln \eta_{1i} &= \sum_{n=1}^N \sum_{k=1}^K (r_{nk} \mathbb{E}_\Phi[\ln \phi_{kix_{ni}}]) + \mathbb{E}_\delta [\ln \delta_i] \\
    \ln \eta_{2i} &= \sum_{n=1}^N \sum_{k=1}^K (r_{nk} \ln \phi_{0ix_{ni}}) + \mathbb{E}_\delta [\ln (1 - \delta_i)]
\end{align}
\begin{equation}
\gamma_i \sim \mbox{Bernoulli}(c_i)
\end{equation}
\begin{align}
    q^\ast(\delta_i) &\propto \delta_i^{(c_i + a - 1)} (1- \delta_i)^{(1 - c_i + a - 1)}
\end{align}
\begin{equation}
    \delta_i \sim \mbox{Beta}(c_i + a, 1 - c_i + a)
\end{equation}

All expectations throughout are taken over the variational distributions for each parameter. The general algorithm now involves cycling between estimating the $r_{nk}$ by using the current variational distributions $q^\ast(\pi)$, $q^\ast(\phi)$ and the current value of $c_i = \mathbb{E}_\gamma(\gamma_i)$ (when using variable selection) to calculate the expectations in Equation \eqref{rhodef2} (“variational E step”), and then using these to recompute the parameters in the variational distributions for $\pi, \Phi, \gamma, \delta$ ("variational M step"). We initialise using k-modes \citep{Chaturvedi2001}, a method analogous to k-means for categorical data.

\subsection{Summarising and Bayesian model averaging}\label{suppsubsec:modelaverage}

\subsubsection{Variation of information}
The most `representative' clustering ${\bf Z}^\ast$ given the co-clustering matrix \eqref{psmeqn} is not well defined, but Binder introduced a formal definition under a Bayesian decision theory framework \citep{BINDER1978}. Given a loss function $L$, the most `representative' clustering ${\bf Z}^\ast$ should satisfy:
\begin{equation}
    {\bf Z}^\ast = \argmin_{\hat{z}} \mathbb{E} [L(z, \hat{z}) | X] \label{decision}
\end{equation}

One such loss function $L$ which can be used in the framework in Equation \ref{decision} is the variation of information (VoI), constructed using information theory and introduced by Meilǎ \citep{Meil2007} for cluster comparison between clustering solutions $c$ and $c'$. Intuitively, this equation compares the information captured in each clustering individually with the information that is shared between both clusterings; the VoI is therefore small when the information shared by both clustering structures is close to the sum of the individual information captured by each clustering. 

This was first proposed as a loss function by Wade and Gharamani \citep{Wade2018}, who found that VoI has many desirable theoretical properties in the distance space in order to be used as a loss function, and also found a computationally efficient lower bound on the expected loss only depending on the posterior through the co-clustering matrix. We implement VoI via the function \textit{minVI} in the R package \textit{mcclust.ext} and consider the optimisation methods `average' and `complete', where the search space is restricted to clusterings found via hierarchical clustering (with distance metric $1-P$) with either average or complete linkage; computing the lower bound for every possible clustering in the search space is infeasible.

VoI has been found to be successful in finding accurate summary clusterings and correct for the overestimation in the number of clusters in many MCMC simulations \citep{chaumenyetal, Wade2018, Rastelli2017, Lijoi2022}, especially compared to other popular loss functions such as Binder's loss \citep{BINDER1978} and the posterior expected adjusted Rand index \citep{Fritsch2009}. 

\subsubsection{Medvedovic clustering}

An alternative method to find ${\bf Z}^\ast$ is what we refer to as Medvedovic clustering \citep{Medvedovic2004}, where $1 - \textbf{P}$ is used as a distance matrix for agglomerative hierarchical clustering with complete linkage. Complete linkage allows the number of clusters to be determined by cutting the tree at a certain linkage distance $1 - \epsilon$ \citep{Fritsch2009}; Fritsch and Ickstadt use a value of $\epsilon = 0.01$.

This approach has been criticised as being \textit{ad hoc} with little theoretic basis; it cannot be expressed in the decision theory framework in Equation \eqref{decision}, and there is no clear principle to choose where to cut the tree, although Fritsch and Ickstadt proved that using $1 - P_{ij}$ as the distance between two observations does hold certain desirable properties for a distance measure as a topological pseudometric for the space of observations \citep{Fritsch2009}. Nevertheless, Medvedovic clustering has shown good results across many biological scenarios \citep{Crook2019, Rasmussen2009}. Medvedovic clustering is implemented in the R package \textit{mcclust} in the function \textit{medv}.

\subsubsection{Summarising the selected variables}

To identify a final summary set of selected variables, we consider a threshold $\tau$ for the proportion of runs in which a variable is selected in. Given $M$ runs with different initialisations, a variable $j$ is selected if:
\begin{equation}
    \frac{\sum_{m=1}^M \mathds{1}(c_j^{(m)} = 1)}{M} > \tau,
\end{equation}
where $c_j^{(m)}$ is the value of $c_j = \mathbb{E}(\gamma_j)$ in run $m$. We found that in many circumstances, $c_j$ fails to converge exactly to 1 but usually gets extremely close to 1, so in practice, we take variable $j$ to be selected if:
\begin{equation}
    \frac{\sum_{m=1}^M \mathds{1}(c_j^{(m)} > 0.5)}{M} > \tau,
\end{equation}

\subsection{Simulated data - results and analysis}

\subsubsection{VICatMix simulations - results}

\begin{figure}[ht]
    \centering
    \includegraphics[scale=0.57]{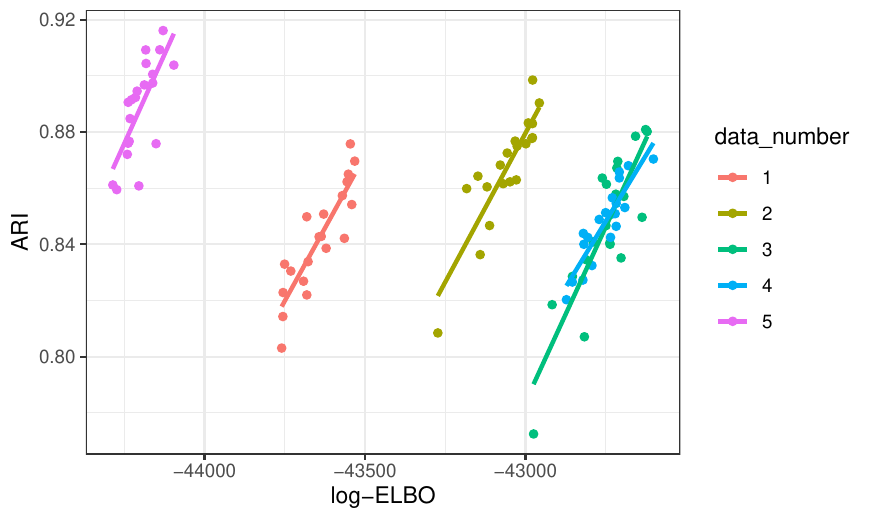}
    \caption{Graph showing the correlation between log-ELBO and ARI across different initialisations of VICatMix on 5 different sets of simulated data.}
    \label{ELBOARIcorr}
\end{figure}

In Figure \ref{ELBOARIcorr}, we look at the correlation between log-ELBO and adjusted Rand Index (ARI) across 5 different batches of simulated data. There was a strong positive correlation between log-ELBO and ARI (with p-values $<$ 0.001 in all cases (Supplementary Table \ref{ELBOARIcorrtable}), justifying picking the clustering with the highest (log-)ELBO as the `optimal' run given many clustering structures with different intialisations. Individual runs of VICatMix performed very well, with an ARI more than 0.80 in almost every initialisation on every dataset. 

\begin{table}[t]
\begin{center}
\caption{Table giving the Pearson correlation coefficient and the p-value of the correlation for each dataset.}\label{ELBOARIcorrtable}
    \begin{tabular}[c]{c | c | c}
    \centering
    Data Number & Correlation Coeff. & p-value \\
    \hline
    1 & 0.880 & 0.00000031 \\
    2 & 0.884 & 0.00000024 \\
    3 & 0.861 & 0.0000011 \\
    4 & 0.896 & 0.000000091 \\
    5 & 0.731 & 0.00025
    \end{tabular}
\end{center}
\end{table}

\begin{figure*}[ht]
    \centering
    \subfloat[4 true clusters, K=10]{
  \includegraphics[scale=0.5]{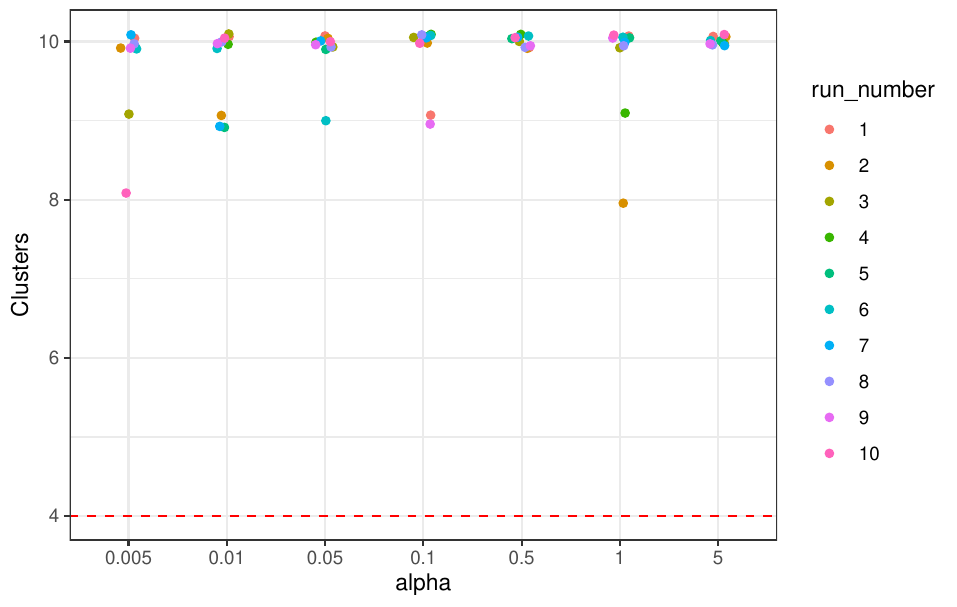}
}
\subfloat[10 true clusters, K=30]{
  \includegraphics[scale=0.5]{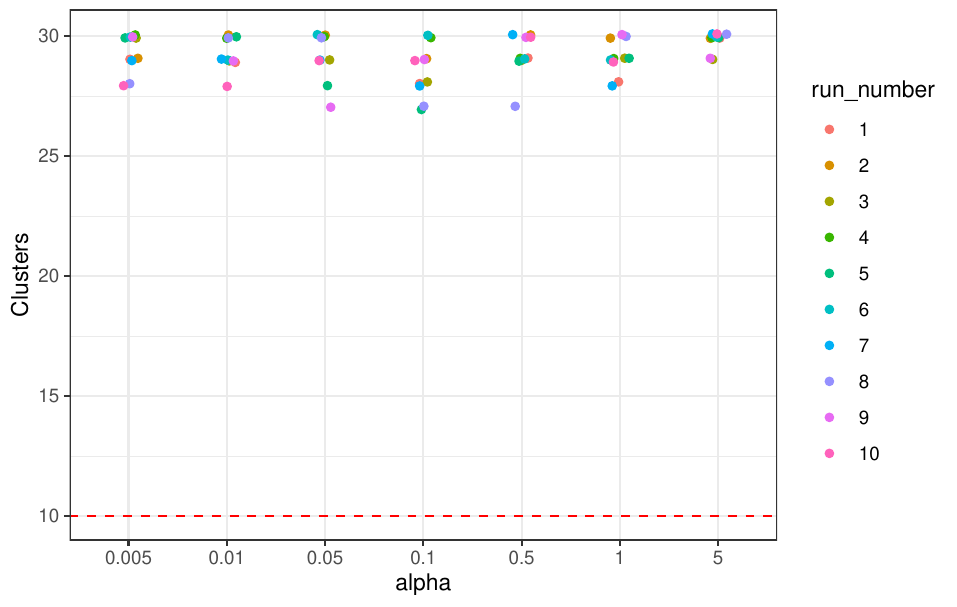}
}
    \caption{Two examples of simulated data with $N = 1000$, $P = 100$ initialised 10 times for each value of $\alpha \in \{0.005, 0.01, 0.05, 0.1, 0.5, 1, 5\}$ with the resulting number of clusters plotted and the `true' number of clusters indicated with a red dashed line. We see that clusters are very rarely emptied.}
    \label{varyingalpha}
\end{figure*}

\begin{figure*}[ht]
    \centering
    \subfloat[4 true clusters, K=10]{
  \includegraphics[scale=0.5]{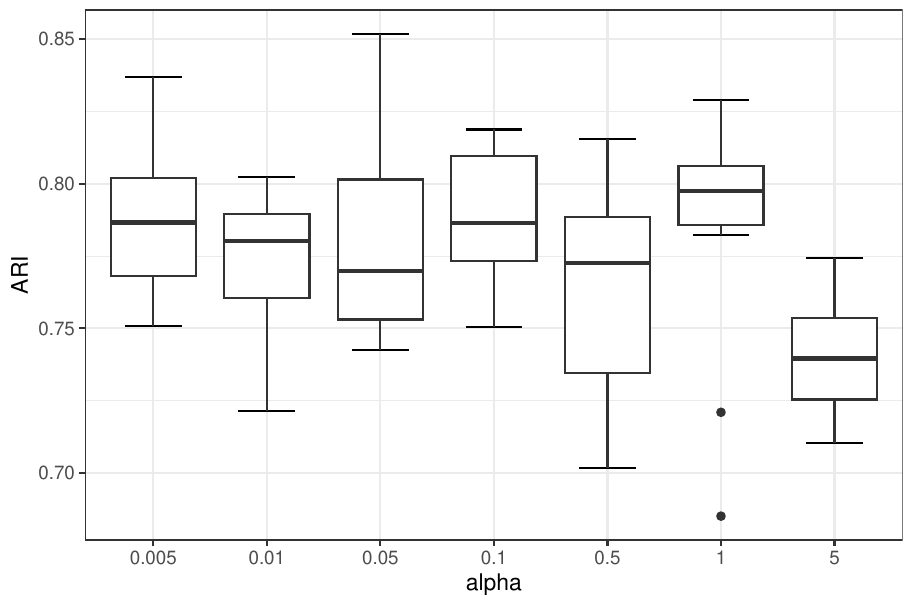}
}
\subfloat[10 true clusters, K=30]{
  \includegraphics[scale=0.5]{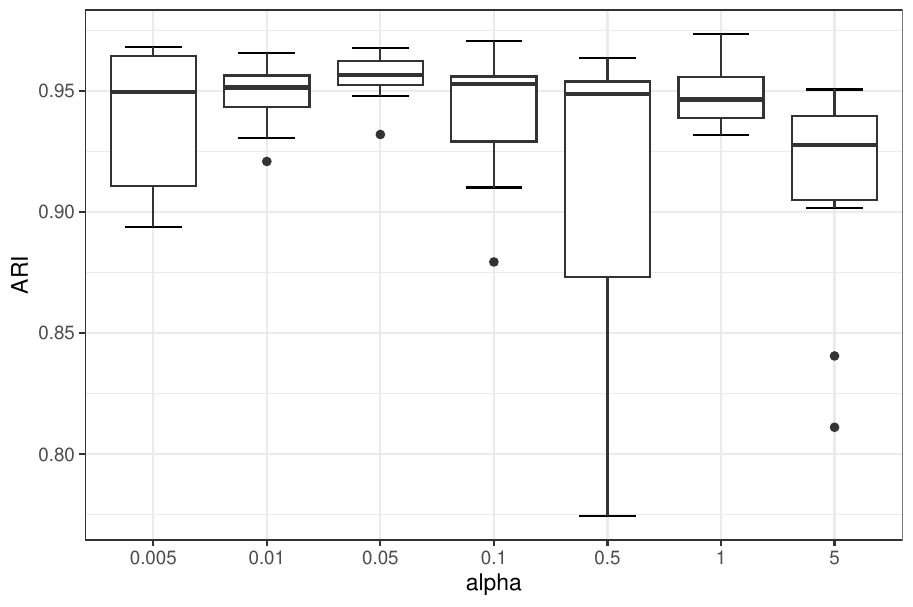}
}
    \caption{Two examples of simulated data with $N = 1000$, $P = 100$ initialised 10 times for each value of $\alpha \in \{0.005, 0.01, 0.05, 0.1, 0.5, 1, 5\}$ with the resulting ARI plotted.}
    \label{varyingalpha2}
\end{figure*}

In Figure \ref{varyingalpha}, we saw that when we varied $\alpha$, the number of clusters in the final converged model still remained close to the initialised value of $K$, despite using an overfitted mixture model. Even if the true posterior behaviour leads to the emptying of negligible clusters, our model only gives us a local optimum where observations can still be stuck in small clusters of less than 5 observations. This motivates the use of summarisation and Bayesian model averaging as described in Section \ref{subsec:modelaverage}. Figure \ref{varyingalpha2} shows that varying $\alpha$ did not have an obvious effect on the ARI.

\begin{figure}[ht]
\centering
\includegraphics[scale=0.58]{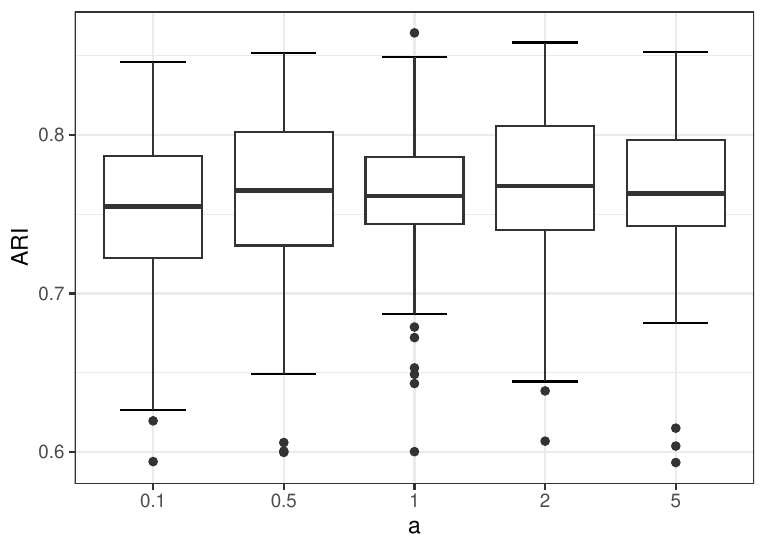}
\caption{Simulated data with $N = 1000$, 75 relevant and 25 irrelevant variables initialised 10 times on 10 different independent datasets for each value of $a \in \{0.1, 0.5, 1, 2, 5\}$ with the resulting distribution of the ARI plotted as a boxplot.}
\label{cexp_ARI}
\end{figure}

Figure \ref{cexp_ARI} demonstrates the effects of changing the hyperparameter in the hyperprior for $\delta$, where $a$ represents the parameter in the Beta distribution. Visually, there appears to be no difference in the accuracy of the model when we change this hyperparameter, and a Kruskal-Wallis test (a non-parametric test to compare the means of the groups) gives a p-value of 0.1598, suggesting that there is no significant difference in the distributions of the groups. We use $a = 2$ as a default throughout the experiments in the report.

\subsubsection{VICatMix-Avg simulations - additional figures}

\begin{table*}[ht]
\begin{center}
\caption{Table comparing the mean ARI of each summarisation method with the grand mean ARI of the runs considered in each size of co-clustering matrix (across all 10 datasets) in Simulation 2.1}
\label{meanARISim1}
\begin{tabular}{c | c | c | c | c}
No. of Runs & Grand Mean ARI & Medv ARI & VoIavg ARI & VoIcomp ARI \\
\hline
5 & 0.815 & 0.892 & \textbf{0.934} & 0.892 \\
10 & 0.842 & 0.924 & \textbf{0.937} & 0.924 \\
15 & 0.826 & 0.934 & \textbf{0.940} & 0.934 \\
20 & 0.836 & 0.940 & \textbf{0.941} & 0.942 \\
25 & 0.841 & 0.939 & \textbf{0.942} & 0.940 \\
30 & 0.845 & 0.940 & \textbf{0.943} & 0.940 \\
\end{tabular}
\end{center}
\end{table*}

\begin{table*}[ht]
\begin{center}
\caption{Table comparing the mean ARI of each summarisation method with the grand mean ARI of the runs considered in each size of co-clustering matrix (across all 10 datasets) in Simulation 2.2}
\label{meanARISim2}
\begin{tabular}{c | c | c | c | c}
No. of Runs & Grand Mean ARI & Medv ARI & VoIavg ARI & VoIcomp ARI \\
\hline
5 & 0.870 & 0.910 & \textbf{0.945} & 0.910 \\
10 & 0.871 & 0.929 & \textbf{0.949} & 0.929 \\
15 & 0.874 & 0.941 & \textbf{0.951} & 0.941 \\
20 & 0.873 & 0.943 & \textbf{0.951} & 0.943 \\
25 & 0.874 & 0.949 & \textbf{0.950} & 0.949 \\
30 & 0.875 & 0.949 & \textbf{0.951} & 0.949 \\
\end{tabular}
\end{center}
\end{table*}

\begin{table*}[ht]
\begin{center}
\caption{Table comparing the mean ARI of each summarisation method with the grand mean ARI of the runs considered in each size of co-clustering matrix (across all 10 datasets) in Simulation 3}
\label{meanARIsSim3}
\begin{tabular}{c | c | c | c | c}
No. of Runs & Grand Mean ARI & Medv ARI & VoIavg ARI & VoIcomp ARI \\
\hline
5 & 0.857 & 0.893 & \textbf{0.904} & 0.893 \\
10 & 0.854 & 0.901 & \textbf{0.906} & 0.901 \\
15 & 0.855 & \textbf{0.907} & 0.907 & \textbf{0.907} \\
20 & 0.854 & \textbf{0.910} & 0.908 & \textbf{0.910} \\
25 & 0.854 & \textbf{0.909} & 0.908 & \textbf{0.909} \\
30 & 0.854 & \textbf{0.910} & 0.907 & \textbf{0.910} \\
\end{tabular}
\end{center}
\end{table*}

\begin{figure*}[ht]
    \centering
    \subfloat[]{\includegraphics[scale=0.25]{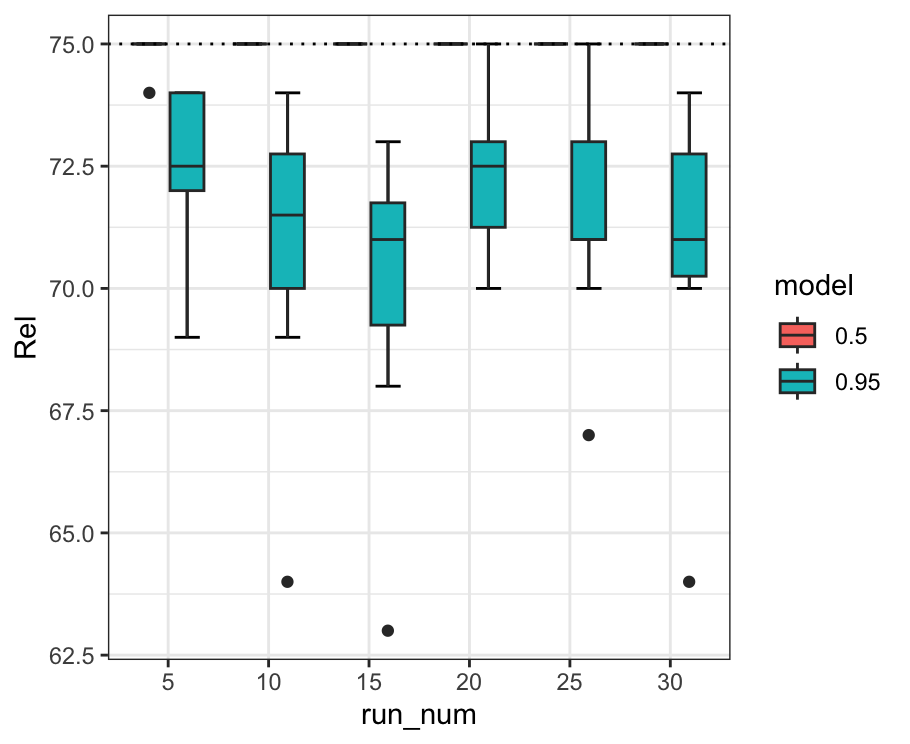}}
    \subfloat[]{\includegraphics[scale=0.25]{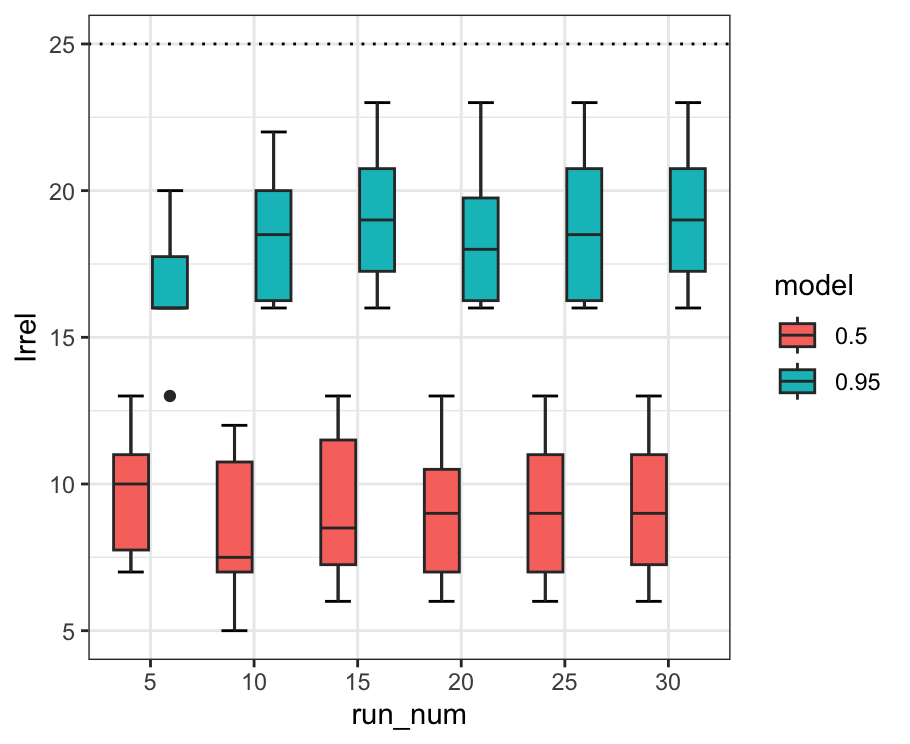}}
    \caption{Boxplots comparing the number of relevant and irrelevant variables discovered by thresholds $\tau = 0.5$ and $\tau = 0.95$ for finding the selected variables in Simulation 2.4 with variable selection. Note that $\tau = 0.5$ almost always finds the correct number of relevant variables, 75.}
    \label{Sim1ModelAvgRelIrrel}
\end{figure*}

\begin{figure*}[ht]
    \centering
    \subfloat[]{\includegraphics[scale=0.25]{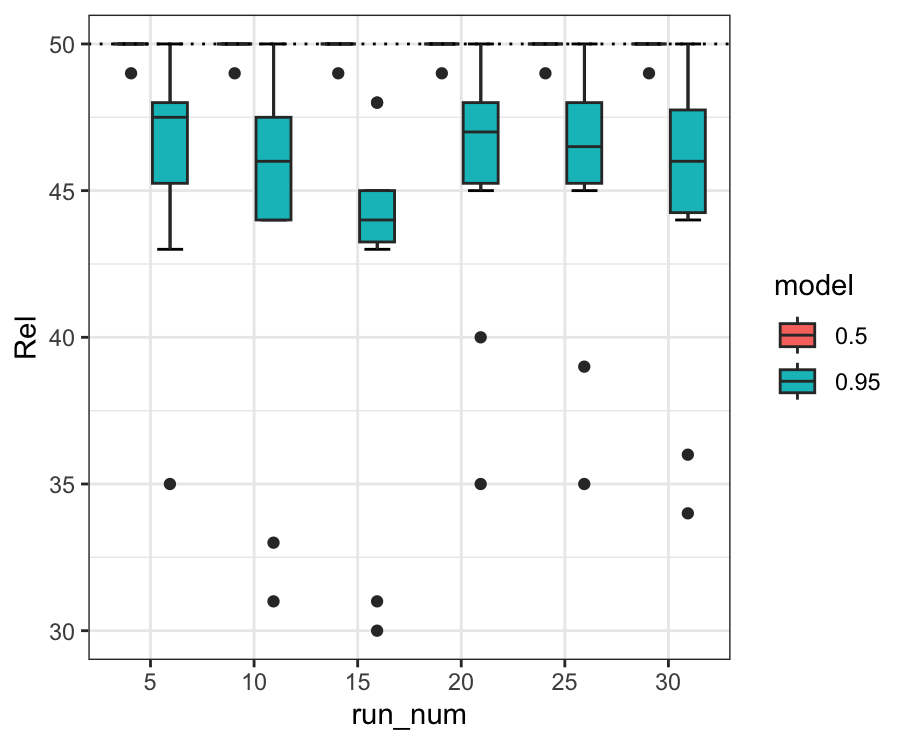}}
    \subfloat[]{\includegraphics[scale=0.25]{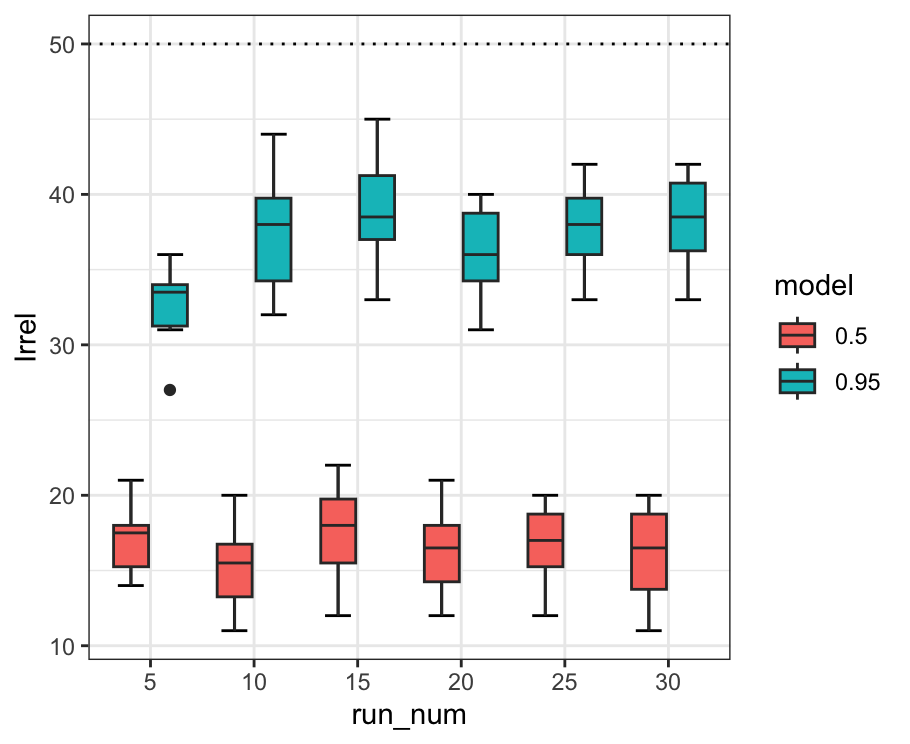}}
    \caption{Boxplots comparing the number of relevant and irrelevant variables discovered by thresholds $\tau = 0.5$ and $\tau = 0.95$ for finding the selected variables in Simulation 2.4 with variable selection. Note that $\tau = 0.5$ almost always finds the correct number of relevant variables, 50.}
    \label{Sim2ModelAvgRelIrrel}
\end{figure*}

We additionally look at the effects of randomising which clustering solutions we use for VoI with complete linkage using 25 runs, selected from 50 different potential runs for $N = 1000$ and $P = 100$, in order to show our solution is robust to the choice of initialisations. We test this on 10 different datasets with 10 different model-averaged solutions for each dataset, and we see from Table \ref{whichrunsinpsm} that the performance of our model-averaged clustering solution is robust to the choice of which runs are selected. All model-averaging solutions had between 10-12 clusters.

\begin{table*}[ht]
\begin{center}
\caption{Table comparing the mean and variance of the ARI of the model-averaged VoI runs for each simulated dataset, with the grand mean of the mean of the individual runs considered for each dataset.}
\label{whichrunsinpsm}
\begin{tabular}{c | c | c | c}
Dataset & Mean Model-Averaged ARI & Variance Model-Averaged ARI & Grand Mean Individual ARI \\
\hline
1 & 0.941 & 0.000043 & 0.864\\
2 & 0.924 & 0.0000089 & 0.832\\
3 & 0.956 & 0.000016 & 0.854\\
4 & 0.954 & 0.0000010 & 0.866\\
5 & 0.934 & 0.0000078 & 0.852\\
6 & 0.934 & 0.000029 & 0.852\\
7 & 0.938 & 0.000022 & 0.864\\
8 & 0.961 & 0.000031 & 0.859\\
9 & 0.905 & 0.000021 & 0.793\\
10 & 0.932 & 0.000021 & 0.844\\
\end{tabular}
\end{center}
\end{table*}

\subsubsection{Comparisons to other models - set-up}
Further details for the comparator methods to VICatMix, including implementation settings, are as follows:

\begin{itemize}
\item PReMiuM \citep{Liverani2015}: a Dirichlet process mixture model allowing for infinite components trained using Gibbs `slice-sampler' MCMC. We use 1000 burn-in samples and 2500 sweeps. We use `partitioning by medioids' to post-process the MCMC samples (the default value), which uses the k-medioids method (a generalisation of k-means) with the dissimilarity matrix as the distance measure; this is relatively inefficient in comparison to other post-processing methods. PReMiuM has options for two types of variable selection, based on approaches proposed by  \citeauthor{Papathomas2012}; binary variable selection, a small modification on the approach used by Chung and Dunson \citep{Chung2009} which uses a cluster specific variable selection approach by associating a binary random variable determining whether covariate j is important to mixture component c, and continuous variable selection, which performs variable selection by associating each covariate with a latent variable taking values between [0, 1] informing whether the covariate is important in supporting a mixture distribution. The continuous variable selection is similar to our method, but our latent variable is a binary random variable rather than a continuous random variable.

\item Bayesian Hierarchical Clustering (BHC) \citep{Heller2005}: implemented via R/Bioconductor \citep{Savage2009}. Performs bottom-up hierarchical clustering, where at each iteration, Bayesian hypothesis testing is used to consider which clusters should be merged. Can be interpreted as an approximate inference method for a Dirichlet process mixture model. There are no user-defined settings for the implementation of BHC in R/Bioconductor, but we note that by default, BHC performs hyperparameter tuning.

\item BayesBinMix \citep{RJ-2017-022}: an MCMC implementation (with tempered MCMC chains to accelerate convergence) for Bayesian finite mixture models; the framework is almost identical to our mixture model, but a discrete prior over $1:K_{max}$ is used to determine the number of clusters. We use 6 heated chains with heats $\{1, 0.92, 0.84, 0.76, 0.68, 0.6\}$, 500 burn-in and 2500 total MCMC sweeps, and a Poisson prior for the number of clusters per the authors' recommendations. BayesBinMix uses two algorithms - `ECR' \citep{Papastamoulis2010} and `KL' \citep{Stephens2000, Rodrguez2014} - to bypass the label switching issue in the MCMC output. The two algorithms give almost identical clustering solutions, although ECR is more efficient; we use ECR in our comparison.

\item FlexMix \citep{Leisch2004}: implements an EM-algorithm under a maximum-likelihood framework for the training of finite mixture models. The \textit{stepFlexmix} function allows for model selection by fitting a given number of models with different initialisations (we use 10) with $k = 1:K$ clusters and finding the maximum likelihood model for each number $k$ of clusters. The overall optimal model with the optimal number of clusters is found using penalised likelihood criteria such as the BIC or ICL criterion \citep{Schwarz1978, Biernacki2000}; we found these gave the same results in all our simulations. 

\item Agglomerative hierarchical clustering: implemented via \textit{hclust} in the R \textit{stats} package with complete linkage \citep{Kaufman1990, Langfelder2012-jq}. We use the average silhouette function implemented in the R package \textit{factoextra} to determine where the tree should be cut.

\end{itemize}

\subsubsection{Comparisons to other models - results and analysis}\label{modelcompareresults}

\begin{figure*}[ht]
    \centering
    \subfloat[]{\includegraphics[scale=0.43]{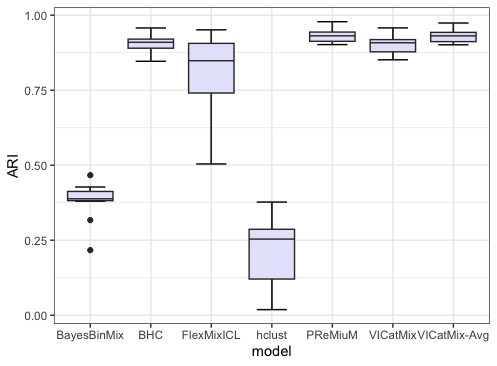}}
    \subfloat[]{\includegraphics[scale=0.43]{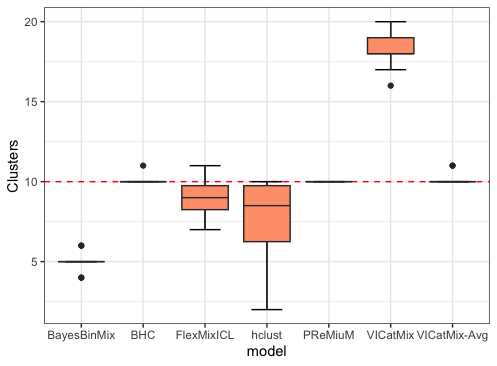}}
    \caption{Comparison of ARI and number of clusters found by each of the methods tested for Simulation 3.1.}
    \label{Sim1NVS}
\end{figure*}

\begin{figure*}[ht]
    \centering
    \subfloat[]{\includegraphics[scale=0.43]{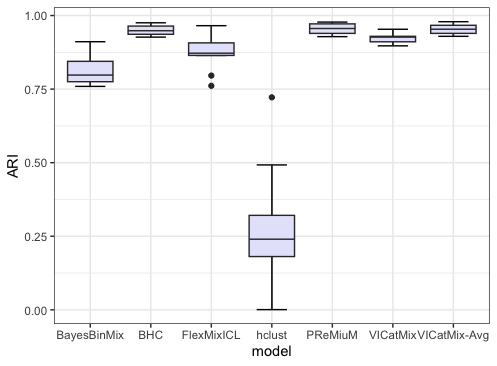}}
    \subfloat[]{\includegraphics[scale=0.43]{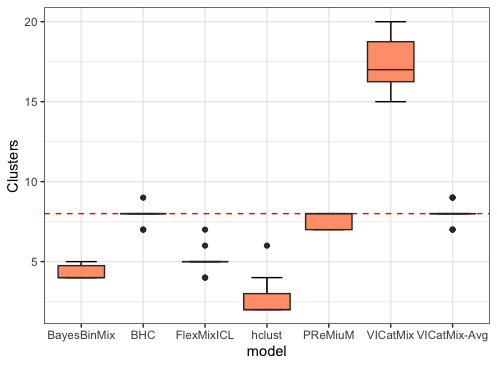}}
    \caption{Comparison of ARI and number of clusters found by each of the methods tested for Simulation 3.2.}
    \label{Sim2NVS}
\end{figure*}

\begin{figure*}[ht]
    \centering
    \subfloat[]{\includegraphics[scale=0.43]{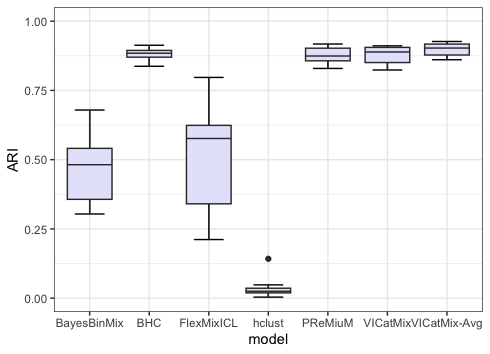}}
    \subfloat[]{\includegraphics[scale=0.43]{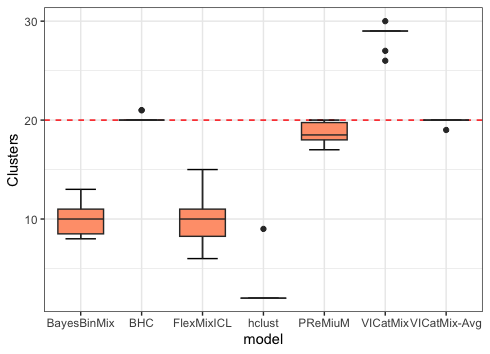}}
    \caption{Comparison of ARI and number of clusters found by each of the methods tested for Simulation 3.3.}
    \label{Sim3NVS}
\end{figure*}

In these simulations, we look the ARI and the the number of clusters chosen by each model, as well as comparing run times. The results for Simulations 3.1, 3.2 and 3.3 are shown in Figures \ref{Sim1NVS}, \ref{Sim2NVS} and \ref{Sim3NVS}, and Table \ref{SimsNVSTime}.

\begin{table*}[ht]
\begin{center}
\caption{Table comparing median and quantiles of run times in seconds for each model in all simulations with no variable selection.}
\label{SimsNVSTime}
\begin{tabular}{c | c | c | c}
Model & Simulation 3.1 & Simulation 3.2 & Simulation 3.3 \\
\hline
BHC & 678.0 [560.3, 897.9] & 653.8 [487.8, 1054.7] & 3481.4 [2908.8, 3838.6] \\
BayesBinMix & 24289.0 [24088.2, 24507.6] & 19958.8 [19646.9, 20150.2] & 53667.0 [53388.4, 54208.2]\\
FlexMix & 92.3 [72.5, 121.6] & 137.4 [97.5, 186.3] & 559.1 [380.1, 1016.5]\\
PReMiuM & 266.0 [259.2, 266.9] & 175.3 [167.1, 244.3] & 1104.5 [1080.5, 1127.1]\\
VICatMix & \textbf{92.0 [83.1, 98.6]} & \textbf{84.2 [64.5, 101.3]} & \textbf{279.7 [140.7, 343.7]}\\
\end{tabular}
\end{center}
\end{table*}

We see that in general, VICatMix-Avg was one of the best-performing models in terms of accuracy across all simulations without noisy variables, achieving ARI scores of more than 0.9 in many cases, and almost always found the correct number of clusters. VICatMix was also more time-efficient, especially as the dataset increased in size. Although individual runs of VICatMix consistently overestimated the true number of clusters as found in Section \ref{averagingresults}, they still outperform other methods in ARI. hclust performed very poorly, BayesBinMix was lengthy and often underestimates the true number of clusters in the data, and FlexMix worked well for smaller datasets but its performance was depleted as we increased the size of the dataset and the number of true clusters in Simulation 3.3, which was also observed by the authors of BayesBinMix \citep{RJ-2017-022}. PReMiuM was the best performing alternative method to VICatMix-Avg, but being an MCMC implementation, it was slower.

Notably, BHC performed well, but much of the computational burden of BHC came from its optimisation of hyperparameters \citep{Heller2005}. Therefore it is possible that by fixing the hyperparameters to pre-determined values, BHC could be competitive in terms of computational time to our variational algorithm; we could also optimise the equivalent hyperparameters in our model and improve accuracy at the cost of time efficiency.

\begin{figure*}[ht]
    \centering
    \subfloat[]{\includegraphics[scale=0.43]{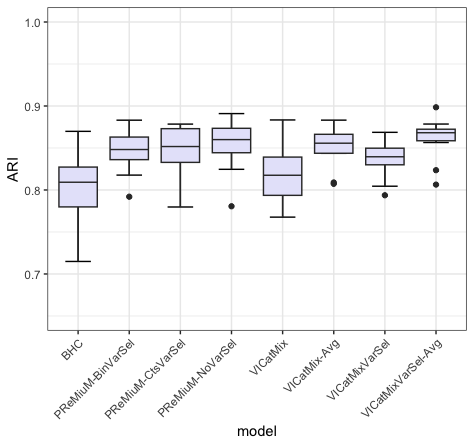}}
    \subfloat[]{\includegraphics[scale=0.43]{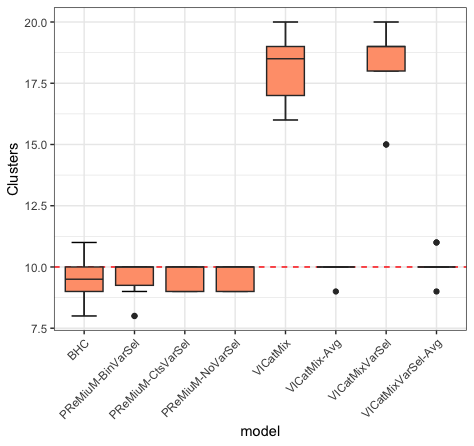}}
    \caption{Comparison of ARI and number of clusters found by each of the methods tested for Simulation 3.4.}
    \label{Sim1VS}
\end{figure*}

\begin{figure*}[ht]
    \centering
    \subfloat[]{\includegraphics[scale=0.43]{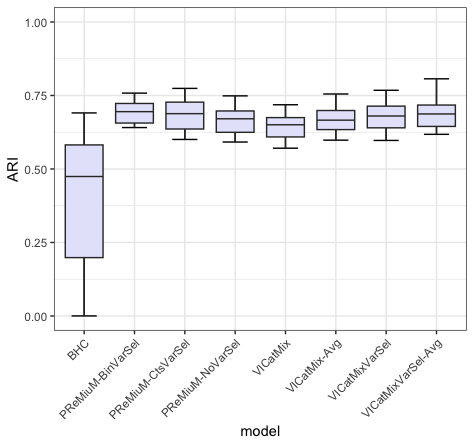}}
    \subfloat[]{\includegraphics[scale=0.43]{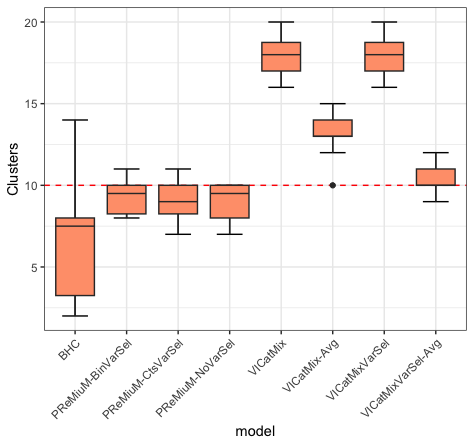}}
    \caption{Comparison of ARI and number of clusters found by each of the methods tested for Simulation 3.5.}
    \label{Sim2VS}
\end{figure*}

Results comparing the ARI and number of clusters for Simulations 3.4 and 3.5 - the noisier datasets - are shown in Figures \ref{Sim1VS} and \ref{Sim2VS}. We see that VICatMixVarSel provided a slight increase in accuracy in terms of ARI and finding the correct number of clusters compared to VICatMix without variable selection and PReMiuM.

In Simulations 3.4 and 3.5, we additionally compared the number of relevant and irrelevant variables correctly identified by the variable selection methods using $F_1$ scores. For VICatMixVarSel, we used a $\tau = 0.95$ variable selection threshold. For a variable to be considered relevant in the PReMiuM models, we looked at a threshold of 0.5 and above for the `rho median', where `rho' relates to the variable selection latent variable in PReMiuM taking values in [0,1] and the `rho median' takes the median across all MCMC sweeps. We also considered a threshold of 0.95 for the rho median, analogous to our own variable selection threshold; this performed poorly compared to 0.5 so is omitted.

Table \ref{f1scores2} shows that variable selection in VICatMix achieved the highest $F_1$ score in Simulation 3.4, and performed similarly to both PReMiuM variable selection methods in Simulation 3.5. Run times are seen in Table \ref{SimsVSTime}.

\begin{table*}[ht]
\begin{center}
\caption{Table comparing median and quantiles of run times in seconds for each model in all simulations with variable selection. For VICatMix, only the times for the optimal runs are used.}
\label{SimsVSTime}
\begin{tabular}{c | c | c }
Model & Simulation 3.4 & Simulation 3.5 \\
\hline
BHC & 417.0 [389.2, 418.7] & 1895.8 [1762.4, 1914.3] \\
PReMiuM-BinVarSel & 294.9 [287.1, 296.9] & 696.7 [676.7, 737.1] \\
PReMIuM-CtsVarSel & 273.8 [263.9, 277.8] & 673.5 [644.1, 733.7]\\
PReMiuM-NoVarSel & 184.7 [179.1, 187.3] & 516.2 [472.8, 547.2] \\
VICatMix & \textbf{108.8 [81.0, 173.5]} & \textbf{324.4 [222.5, 393.8]} \\
VICatMix-VarSel & \textbf{193.0 [153.3, 241.7]} & \textbf{627.8 [454.1, 761.9]}\\
\end{tabular}
\end{center}
\end{table*}

\begin{table}[ht]
\begin{center}
\caption{Table comparing mean $F_1$ scores for variable selection methods under both Simulation 1 and Simulation 2. For VICatMixVarSel, we take the mean of the $F_1$ scores for only the optimal (highest ELBO) runs.}
\label{f1scores2}
\begin{tabular}{c | c | c }
Methods & Simulation 1 & Simulation 2 \\
\hline
PReMiuM-BinVarSel & 0.862 & 0.930 \\
PReMiuM-CtsVarSel & 0.951 & \textbf{0.980} \\
VICatMixVarSel & 0.956 & 0.852 \\
VICatMixVarSel-Avg & \textbf{0.969} & 0.931 \\
\end{tabular}
\end{center}
\end{table}

\subsubsection{Run-times}

\begin{figure*}[ht]
    \centering
    \subfloat[No variable selection]{
  \includegraphics[scale=0.2]{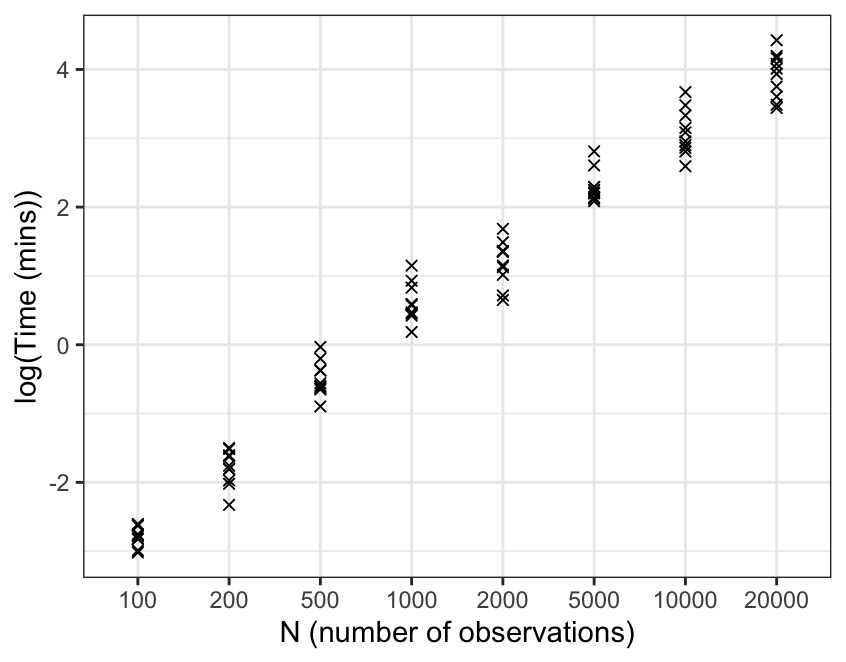}
}
\subfloat[Variable selection]{
  \includegraphics[scale=0.2]{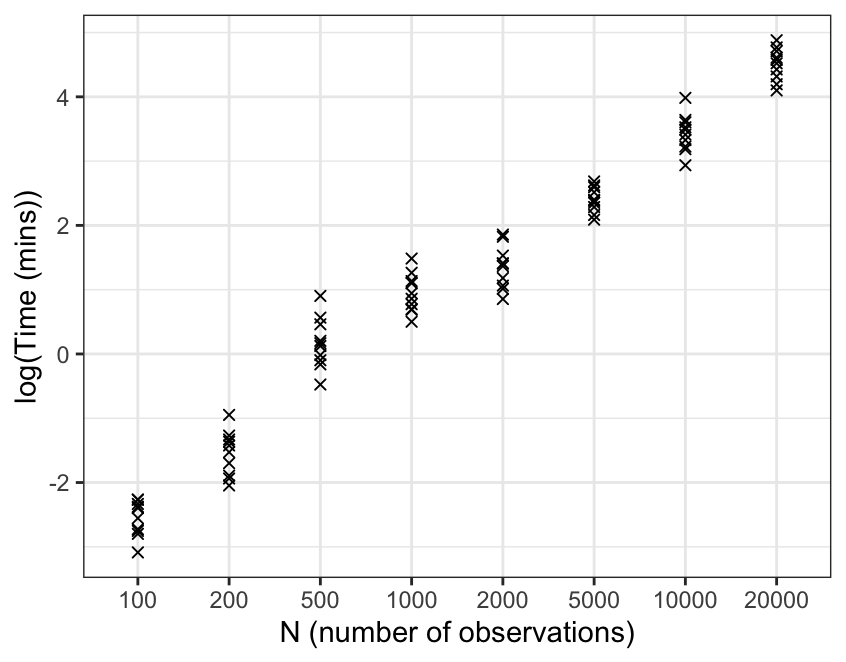}
}
    \caption{Graph showing how run-time of VICatMix varies as we increase $N$, the number of observations in our dataset in our run-times simulation study.}
    \label{runtimes}
\end{figure*}

\begin{figure}[ht]
    \centering
    \includegraphics[scale=0.25]{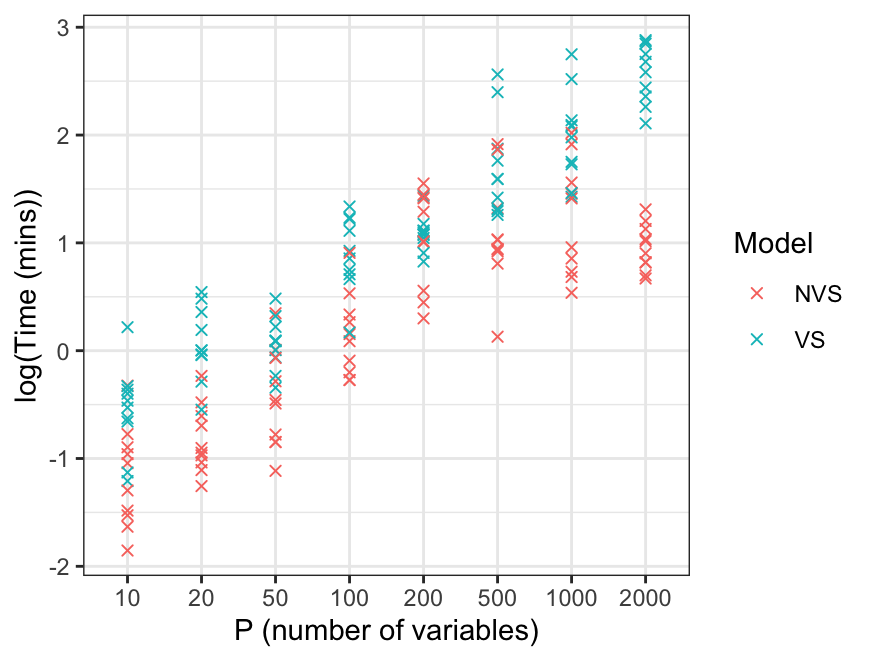}
    \caption{Graph showing how both the time and ARI of VICatMix (with and without variable selection) varies as we increase $P$, the number of covariates in our dataset.}
    \label{runtimesp}
\end{figure}

\begin{figure*}[ht]
    \centering
    \subfloat[Changing N where $P = 100$.]{\includegraphics[scale=0.25]{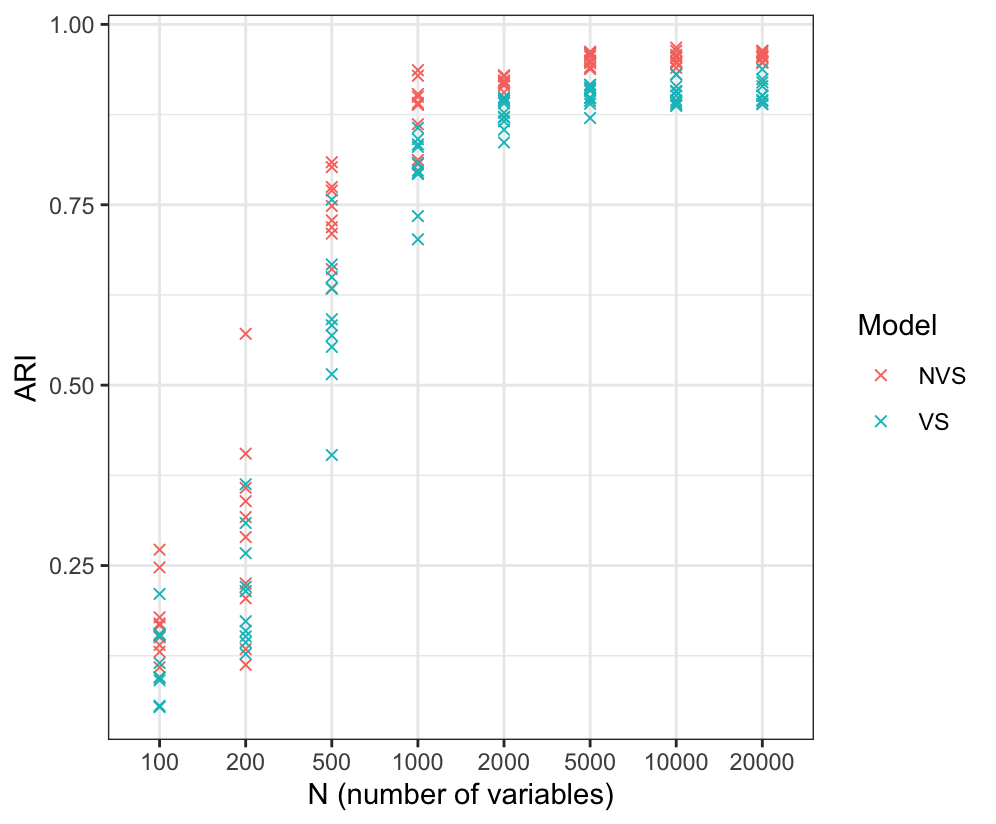}}
    \subfloat[Changing P where $N = 1000$.]{\includegraphics[scale=0.25]{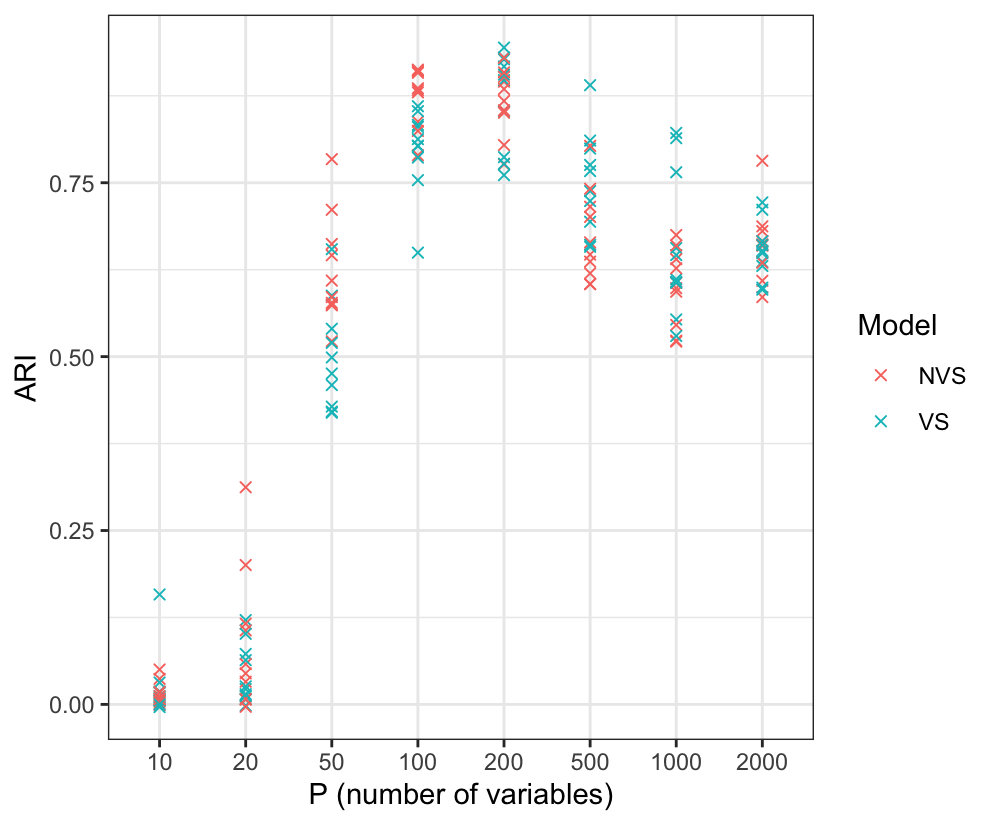}}
    \caption{Graph showing how the ARI of VICatMix varies as we increase $N$ or $P$, the number of observations or the number of variables in the dataset. We generate 10 independent datasets for each value of $N$ or $P$ with and without variable selection with 10 true clusters, and we initialise with 20 clusters. In both cases with variable selection, 80\% of variables are relevant.}
    \label{runtimesARI}
\end{figure*}

We saw in Section \ref{modelcompareresults} that VICatMix was faster than other commonly used models in R. We see in Figures \ref{runtimes} and \ref{runtimesp} that the run-time of our model - both with and without variable selection - approximately scaled linearly with both the number of observations (N) and the number of covariates (P). Running VICatMix (and given sufficient compute, VICatMix-Avg) is feasible for datasets with at least 20000 observations; a run takes between 0.5-2 hours without variable selection and 1-3 hours for variable selection for $N=20000$. Figure \ref{runtimesARI} shows that the accuracy of our model generally improved as $N$ increases, although we also see that the accuracy of our model in terms of ARI was depleted slightly as we increased $P$ to $P \gtrsim N$.

\subsubsection{Categorical simulation study}\label{catsim}

\begin{figure*}[ht]
    \centering
    \subfloat[Data ordered by true clusters]{\includegraphics[scale=0.43]{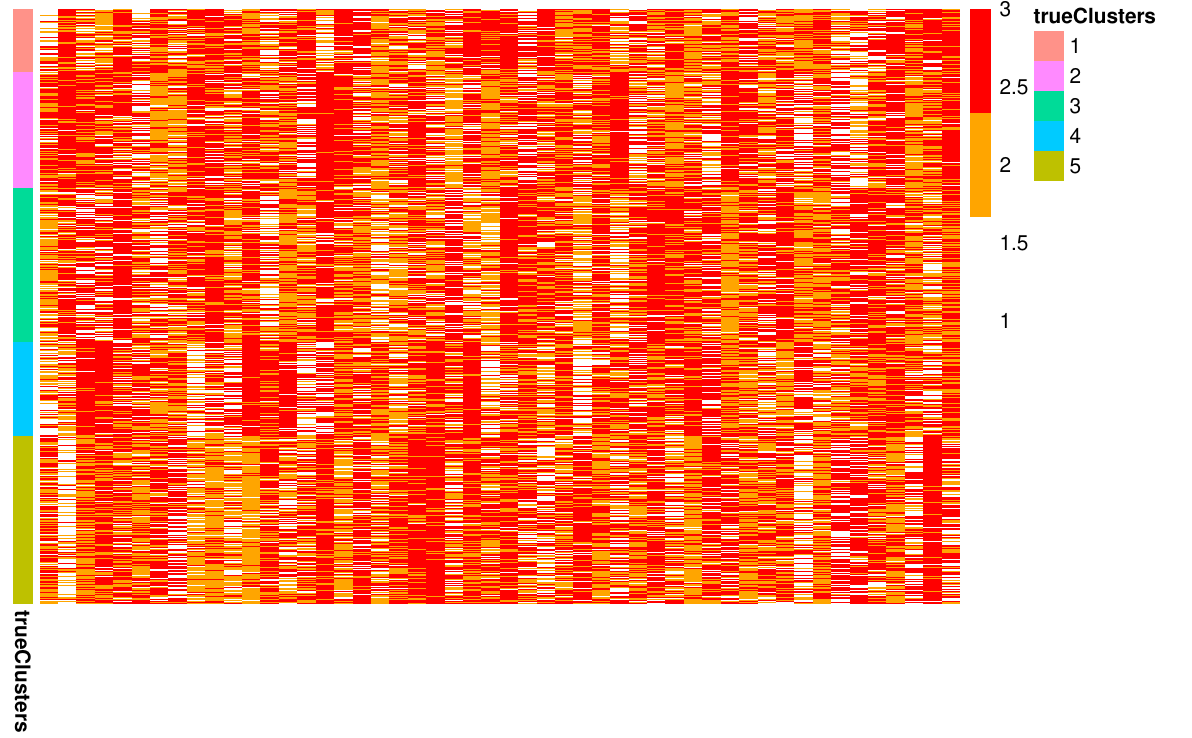}}
    \subfloat[Data ordered by VICatMix clusters]{\includegraphics[scale=0.43]{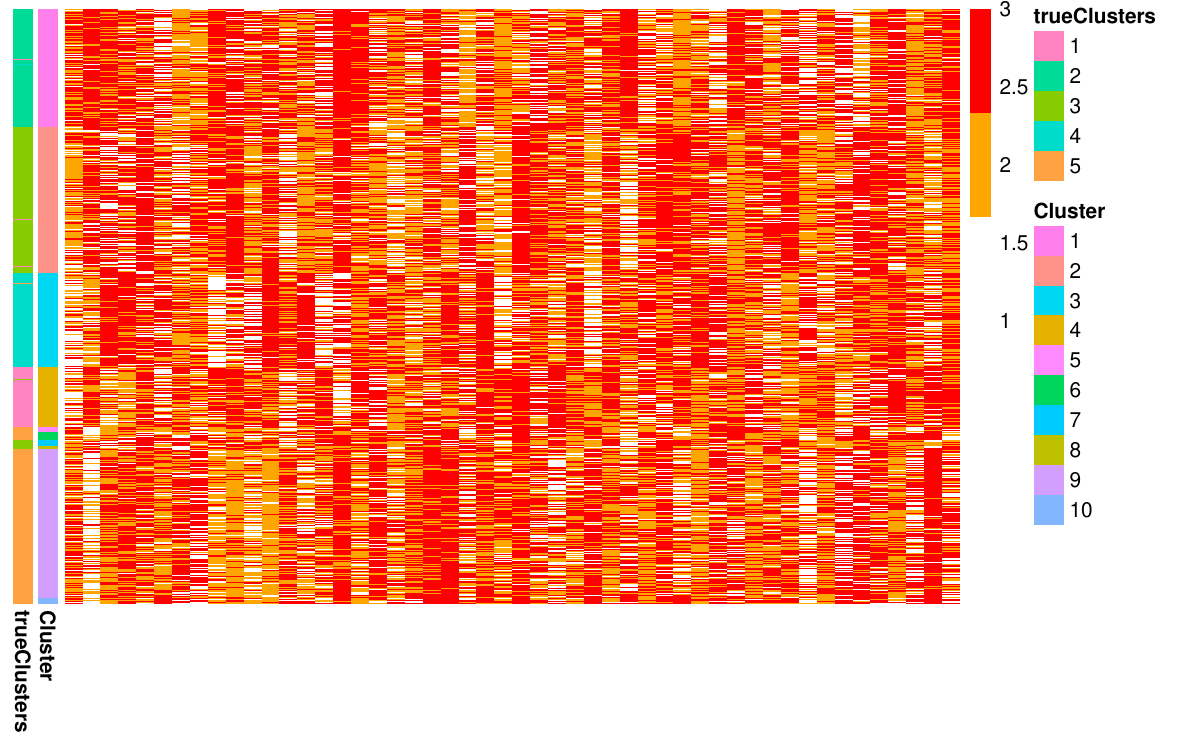}}
    \caption{Heatmaps illustrating the clustering structure of categorical simulated data with $N = 1000$, $P = 50$, 5 true clusters and three categories per variable. VICatMix was run with $\alpha = 0.01$ and $K = 10$ and had an ARI of 0.910 with the true clustering structure.}
    \label{CatHeatmaps}
\end{figure*}

We briefly illustrate that our model can be used with categorical data where variables have more than 2 possible categories in Figure \ref{CatHeatmaps}, where we generated simulated data with 3 categories per variable.

\begin{figure*}[ht]
    \centering
    \subfloat[Data ordered by true clusters]{\includegraphics[scale=0.55]{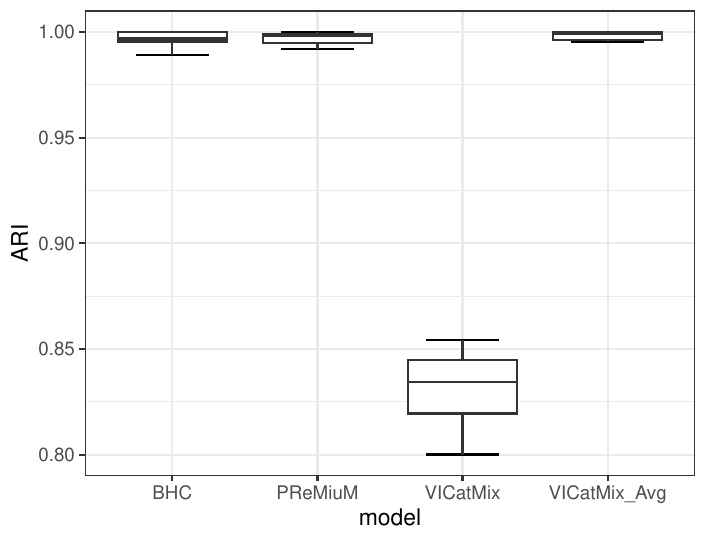}}
    \subfloat[Data ordered by VICatMix clusters]{\includegraphics[scale=0.55]{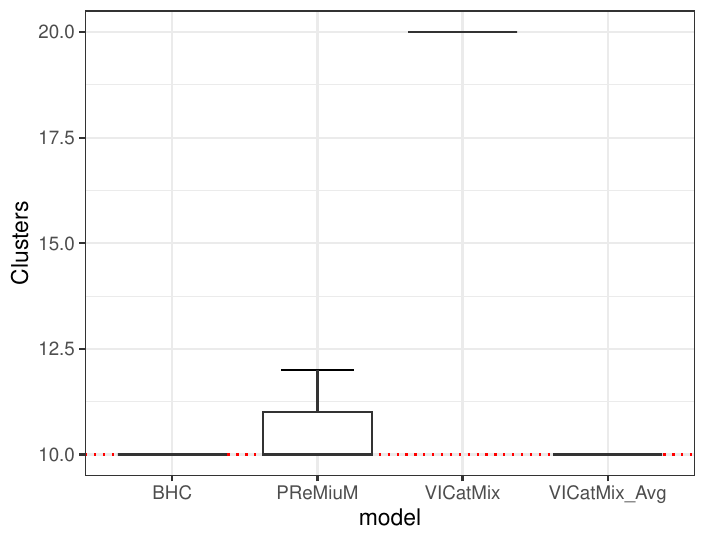}}
    \caption{Boxplots comparing the distribution of ARI and number of clusters on 10 independently generated datasets after running VICatMix, VICatMix-Avg, BHC and PReMiuM on simulated data with $N = 1000$, $P = 100$, 10 evenly sized true clusters and three categories per variable. VICatMix was run with $\alpha = 0.05$ and $K = 20$ and for the VICatMix with no averaging, we took the run with the maximum ELBO.}
    \label{SimCatCompare}
\end{figure*}

We performed a simulation study comparing the performance of VICatMix with categorical data with BHC and PReMiuM. All of VICatMix-Avg, BHC and PReMIuM achieved extremely good results on all datasets, with ARI values between 0.98 and 1 seen in Figure \ref{SimCatCompare}.

\subsection{Yeast galactose data}

\begin{figure}[t]
    \centering
    \includegraphics[scale=0.23]{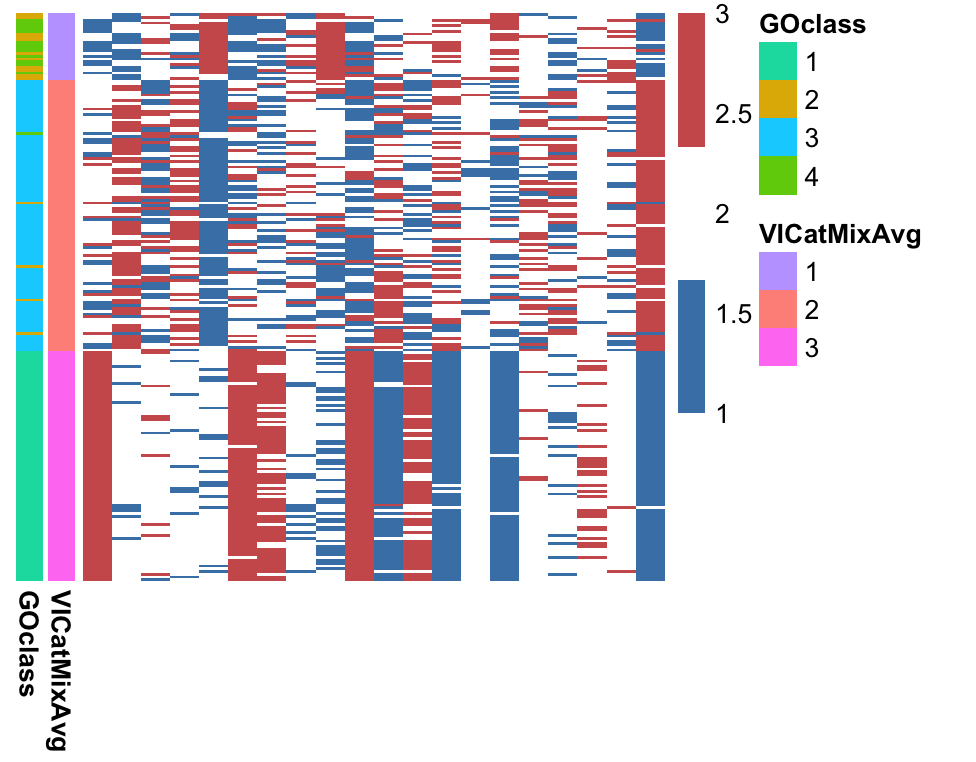}
    \caption{Heatmap of the VICatMix-Avg clustering structure on the yeast galactose data compared with the GO labelling when K=4.}
    \label{Galactose4Heat}
\end{figure}

\subsection{Acute myeloid leukaemia}

\begin{figure}[ht]
    \centering
    \includegraphics[scale=0.23]{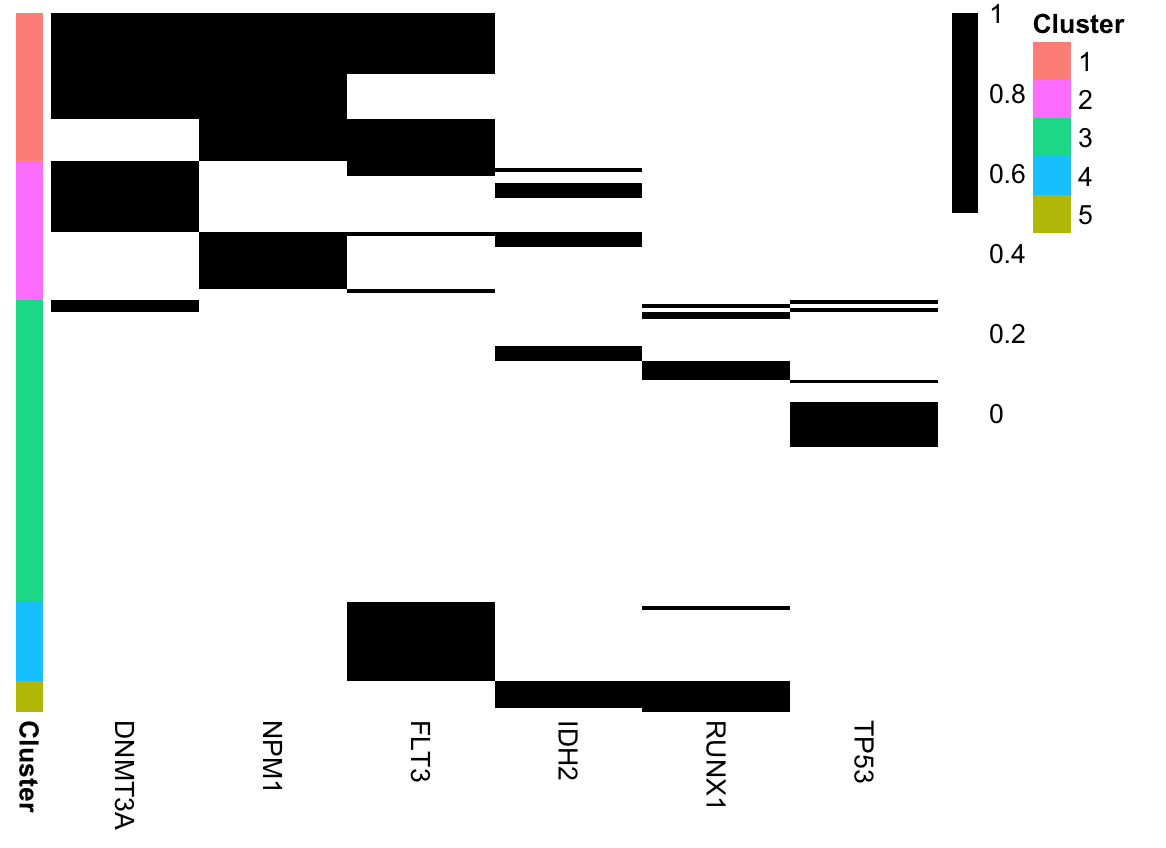}
    \caption{Heatmap of AML with rows reordered.}
    \label{ogAML}
\end{figure}

In Figure \ref{ogAML}, we see that the samples in Cluster 2 had \textit{either} a DNMT3A mutation \textit{or} a NPM1 mutation. This could be indicative of further subclustering structure within Cluster 2; with different parameters for the prior on $\phi$, it is possible that this cluster could have been split.

\subsection{Pan-cancer cluster-of-clusters analysis with K = 40}\label{suppPancan}

\begin{sidewaysfigure*}
\centering
\includegraphics[scale=0.5]{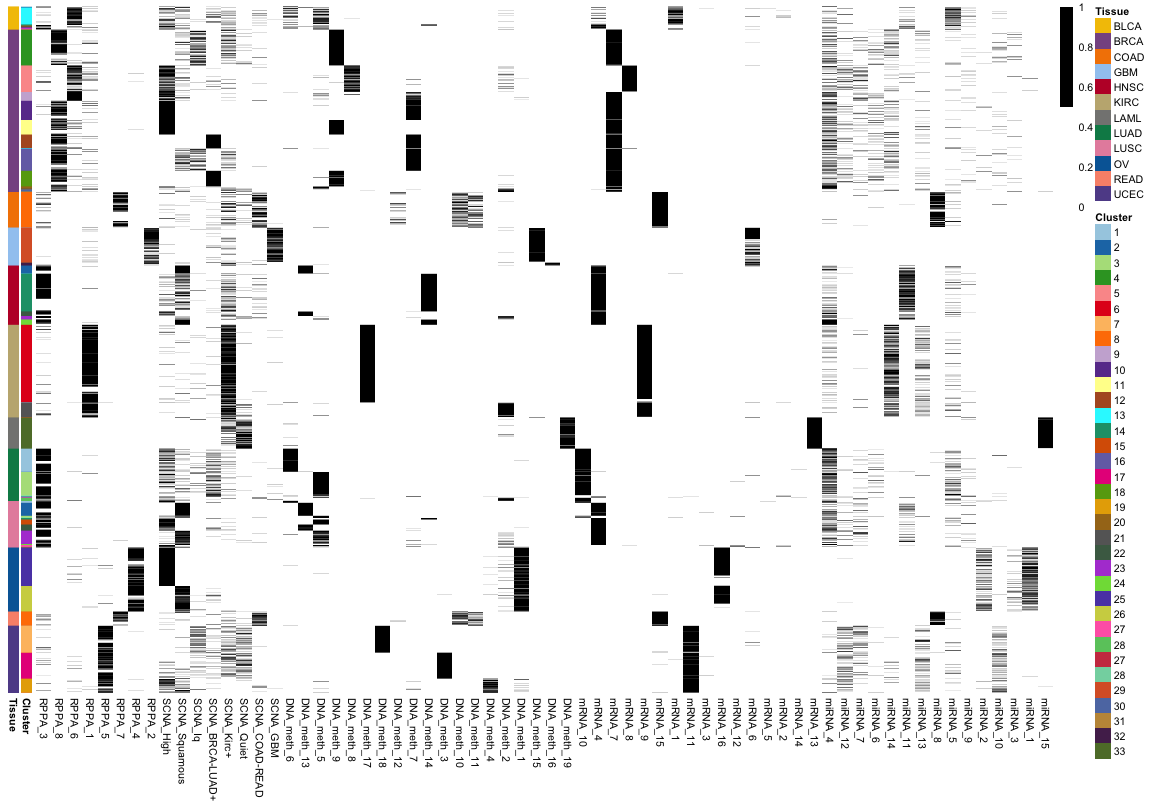}
\caption{A heatmap showing the VICatMix-Avg clustering of the Matrix of Clusters for our pan-cancer data, comparing the clusters to the tissue of origin.}
\label{PheatmapCOCA}
\end{sidewaysfigure*}

\begin{figure}[ht]
\centering
\includegraphics[scale=0.5]{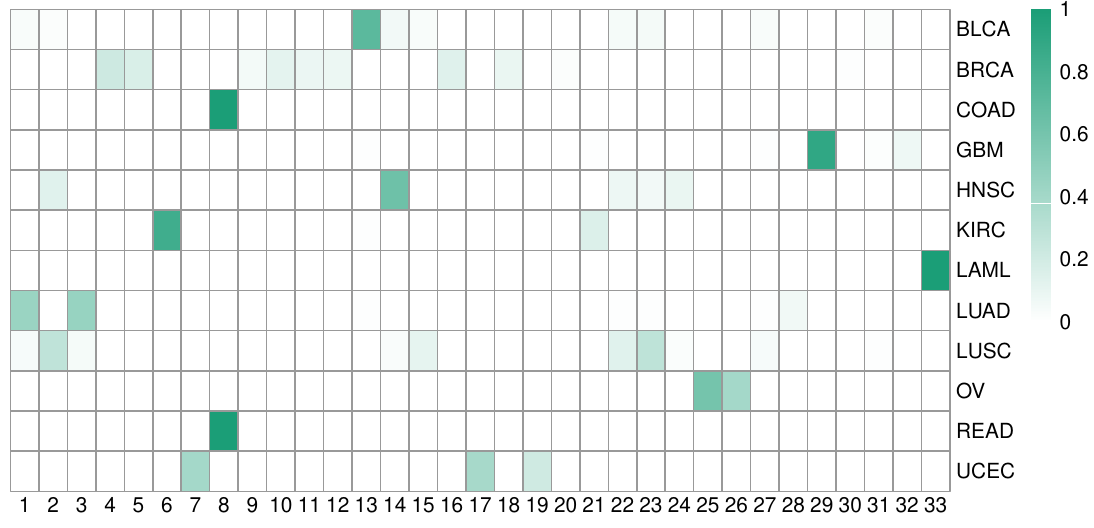}
\caption{A heatmap showing the correspondence between clusters produced by our model and tissues of origin. A darker cell colour in row i indicates a higher percentage of samples from tissue i are in the given cluster j.}
\label{PercentCOCA}
\end{figure}

We illustrate here the performance of our model when considering $K=40$ in each run of VICatMix in order to investigate sub-clustering structures within tissues of origin. We see in Figures \ref{PercentCOCA} and \ref{PheatmapCOCA} that once again the clusters clearly corresponded with tissue of origin, and many tissues were subdivided into multiple clusters. For example, Cluster 33 corresponded exactly with LAML samples; Clusters 7, 17 and 19 corresponded with UCEC samples; and Cluster 8 corresponded precisely with a mixed COAD/READ cluster. Both are colorectal adenocarcinomas and have been previously reported to have very similar expression features at the molecular level - for example, a study by TCGA suggested that non-hypermutated adenocarcinomas of the colon and rectum were not distinguishable at the genomic level \citep{Colorectal2012}, and analysis in the Molecular Epidemiology of Colorectal Cancer (MECC) study \citep{SanzPamplona2011} suggested that there were minimal statistically significant differentially expressed genes between different tumour locations in colorectal cancer patients. 

Notably, when we considered the model in Section \ref{resultsCOCA} with $K=15$ we saw that clusters within tissue types were combined. This suggests that VICatMix is able to detect a hierarchical structure within data; with the number of clusters restricted, clusters were generally separated by tissue type.

\begin{sidewaysfigure*}
\centering
\includegraphics[scale=0.8]{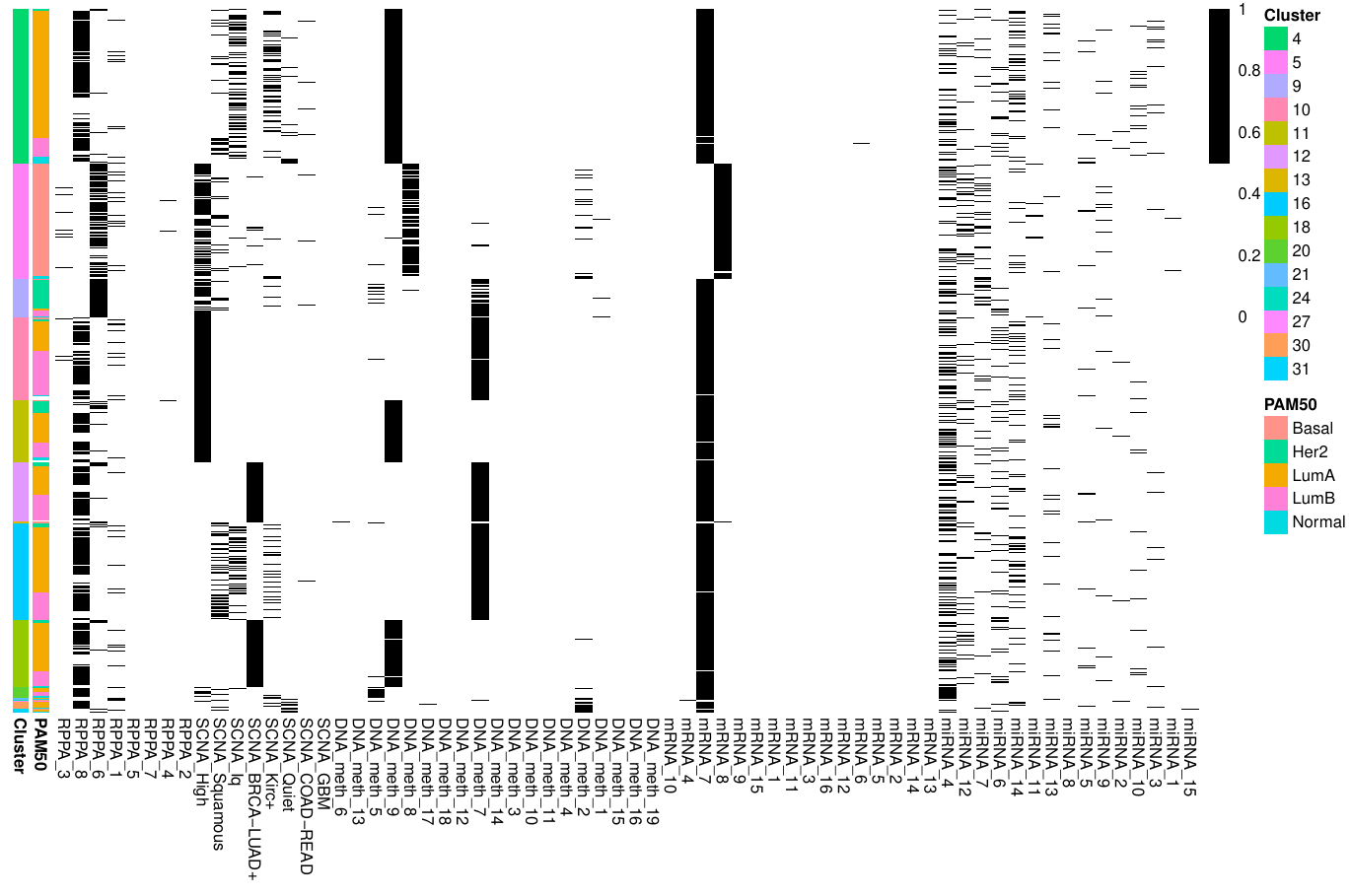}
\caption{A heatmap showing the clustering of all breast cancers in our pan-cancer data, comparing our optimal VICatMix clusters to the PAM50 subtype from \citet{Berger2018}.}
\label{BRCACOCA}
\end{sidewaysfigure*}

Looking at the BRCA samples in depth (Figure \ref{BRCACOCA}), we found that the Basal-like samples were again clearly separated, with 133/141 Basal samples falling into this cluster. Furthermore, Cluster 9 contained over half of the HER2-enriched samples (34/66). This suggests that our clustering method is able to identify and separate another clinically relevant subtype. Other clusters were mostly mixtures of Luminal A and Luminal B samples; it could be of biological interest to look into why these samples were placed in different DNA methylation and somatic copy number clusters. 

VICatMix-Avg's ability to identify these PAM50 BRCA subtypes motivates its application to the identification of other cancer subtypes. For example, ovarian cancer samples were divided into two clusters, seemingly based on somatic copy number clustering where samples are in the SCNA-High and SCNA-Squamous clusters. Serous ovarian carcinomas are known to have extensive copy number alterations (where few are recurrent) and highly complex genomic profiles \citep{Macintyre2018, Hoadley2014}, making it difficult to investigate the mutational processes leading to copy number changes. Little progression has been made in identifying robust clinical subtypes, and gene expression subtypes proposed thus far, such as a 4-subtype classification by the TCGA, have been found to lack robustness across other independent cohorts of patients \citep{Verhaak2012, Chen2018, TCGAOV2011}. We found little correlation between the two clusters and the 4 gene expression subtypes. It could therefore be interesting to investigate what is driving these ovarian cancer samples to fall into different SCNA clusters under the hierarchical clustering analysis by \citeauthor{Hoadley2014} which may motivate the identification of clinically relevant ovarian cancer genomic subtypes.

\begin{table*}[!ht]
    \centering
    \begin{tabular}{|l|l|l|l|l|l|l|l|l|l|l|l|l|l|l|l|}
    \hline
        \textbf{Tissue} & \textbf{1} & \textbf{2} & \textbf{3} & \textbf{4} & \textbf{5} & \textbf{6} & \textbf{7} & \textbf{8} & \textbf{9} & \textbf{10} & \textbf{11} & \textbf{12} & \textbf{13} & \textbf{14} & \textbf{15} \\ \hline
        \textbf{BLCA} & 0 & 21 & 0 & 0 & 0 & 0 & 0 & 0 & 97 & 1 & 0 & 0 & 0 & 1 & 0 \\ \hline
        \textbf{BRCA} & 0 & 1 & 333 & 138 & 1 & 0 & 0 & 316 & 7 & 29 & 0 & 8 & 0 & 1 & 0 \\ \hline
        \textbf{COAD} & 0 & 0 & 0 & 0 & 0 & 0 & 182 & 0 & 0 & 0 & 0 & 0 & 0 & 0 & 0 \\ \hline
        \textbf{GBM} & 0 & 0 & 0 & 0 & 0 & 0 & 0 & 0 & 3 & 0 & 0 & 0 & 189 & 3 & 0 \\ \hline
        \textbf{HNSC} & 0 & 302 & 0 & 0 & 0 & 0 & 0 & 0 & 2 & 0 & 0 & 0 & 0 & 1 & 0 \\ \hline
        \textbf{KIRC} & 0 & 0 & 0 & 0 & 471 & 0 & 0 & 0 & 3 & 0 & 0 & 0 & 1 & 0 & 0 \\ \hline
        \textbf{LAML} & 0 & 0 & 0 & 0 & 0 & 0 & 0 & 0 & 0 & 0 & 0 & 0 & 0 & 0 & 161 \\ \hline
        \textbf{LUAD} & 255 & 6 & 0 & 0 & 0 & 1 & 0 & 0 & 8 & 0 & 0 & 0 & 0 & 0 & 0 \\ \hline
        \textbf{LUSC} & 16 & 206 & 0 & 0 & 0 & 0 & 0 & 0 & 16 & 0 & 0 & 0 & 0 & 0 & 0 \\ \hline
        \textbf{OV} & 0 & 0 & 0 & 0 & 0 & 0 & 0 & 0 & 1 & 0 & 327 & 0 & 0 & 1 & 0 \\ \hline
        \textbf{READ} & 0 & 0 & 0 & 0 & 0 & 0 & 73 & 0 & 0 & 0 & 0 & 0 & 0 & 0 & 0 \\ \hline
        \textbf{UCEC} & 0 & 0 & 0 & 0 & 0 & 342 & 1 & 0 & 2 & 0 & 0 & 0 & 0 & 0 & 0 \\ \hline
    \end{tabular}
    \caption{Cluster assignments by tissue for the pan-cancer data where K=15.}
    \label{FreqtableCOCA15}
\end{table*}

\end{document}